\documentclass[acmlarge, nonacm]{acmart}
\AtBeginDocument{%
  }

\acmJournal{IMWUT}
\usepackage{graphicx}      % 用于插入图片以及使用 \resizebox 自动缩放表格
\usepackage{amsmath}       % 强大的数学公式支持环境
\usepackage{booktabs}      % 顶会标配的优雅三线表 (\toprule, \midrule, \bottomrule)
\usepackage{multirow}      % 允许表格跨列和跨行合并 (\multirow)
\usepackage{subcaption}
\usepackage[table]{xcolor} % 极度重要！支持表格单元格和整行填色 (\rowcolor)
\begin{document}

\title{Towards Green Wearable Computing: A Physics-Aware Spiking Neural Network for Energy-Efficient IMU-based Human Activity Recognition}

\author{Naichuan Zheng}
\email{2022110134zhengnaichuan@bupt.edu.cn}
\affiliation{%
  \department{School of Information and Communication Engineering}
  \institution{Beijing University of Posts and Telecommunications}
  \city{Beijing}
  \country{China}
}

\author{Hailun Xia}
\authornote{Corresponding author.}
\email{xiahailun@bupt.edu.cn}
\affiliation{%
  \department{School of Information and Communication Engineering}
  \institution{Beijing University of Posts and Telecommunications}
  \city{Beijing}
  \country{China}
}

\author{Zepeng Sun}
\email{szp2025140107@bupt.edu.cn}
\affiliation{%
  \department{School of Information and Communication Engineering}
  \institution{Beijing University of Posts and Telecommunications}
  \city{Beijing}
  \country{China}
}

\author{Weiyi Li}
\email{liweiyi@bupt.edu.cn}
\affiliation{%
  \department{International School}
  \institution{Beijing University of Posts and Telecommunications}
  \city{Beijing}
  \country{China}
}

\author{Yingzhe Zhou}
\email{zhouyinzhe@bupt.edu.cn}
\affiliation{%
  \department{School of Information and Communication Engineering}
  \institution{Beijing University of Posts and Telecommunications}
  \city{Beijing}
  \country{China}
}

\renewcommand{\shortauthors}{Zheng et al.}
\renewcommand{\shortauthors}{Trovato et al.}

\begin{abstract}

Wearable IMU-based Human Activity Recognition (HAR) relies heavily on Deep Neural Networks (DNNs), which are burdened by immense computational and buffering demands. Their power-hungry floating-point operations and rigid requirement to process complete temporal windows severely cripple battery-constrained edge devices. While Spiking Neural Networks (SNNs) offer extreme event-driven energy efficiency, standard architectures struggle with complex biomechanical topologies and temporal gradient degradation. To bridge this gap, we propose the Physics-Aware Spiking Neural Network (PAS-Net), a fully multiplier-free architecture explicitly tailored for Green HAR. Spatially, an adaptive symmetric topology mixer enforces human-joint physical constraints. Temporally, an $O(1)$-memory causal neuromodulator yields context-aware dynamic threshold neurons, adapting actively to non-stationary movement rhythms. Furthermore, we leverage a temporal spike error objective to unlock a flexible early-exit mechanism for continuous IMU streams. Evaluated across seven diverse datasets, PAS-Net achieves state-of-the-art accuracy while replacing dense operations with sparse 0.1 pJ integer accumulations. Crucially, its confidence-driven early-exit capability drastically reduces dynamic energy consumption by up to 98\%. PAS-Net establishes a robust, ultra-low-power neuromorphic standard for always-on wearable sensing. The source code and pre-trained models are publicly available at \url{https://github.com/zhengnaichuan2022/PAS-Net.git}.
\end{abstract}
\begin{CCSXML}
<ccs2012>
   <concept>
       <concept_id>10003120.10003138</concept_id>
       <concept_desc>Human-centered computing~Ubiquitous and mobile computing</concept_desc>
       <concept_significance>500</concept_significance>
       </concept>
   <concept>
       <concept_id>10010147.10010257.10010293.10011809</concept_id>
       <concept_desc>Computing methodologies~Bio-inspired approaches</concept_desc>
       <concept_significance>500</concept_significance>
       </concept>
 </ccs2012>
\end{CCSXML}

\ccsdesc[500]{Human-centered computing~Ubiquitous and mobile computing}
\ccsdesc[500]{Computing methodologies~Bio-inspired approaches}

\keywords{Human Activity Recognition, Spiking Neural Networks, Wearable Computing, Energy Efficiency, Sensor Topology, Inertial Sensors}

\maketitle
\section{Introduction}

\label{sec:intro}

Human Activity Recognition (HAR) utilizing wearable Inertial Measurement Units (IMUs) has established itself as a fundamental pillar of ubiquitous computing \cite{bulling2014tutorial,haresamudram2025past}. By continuously monitoring and interpreting body movements, IMU-based HAR empowers a myriad of critical applications, ranging from pervasive healthcare \cite{patel2012review} and eldercare monitoring to professional sports analytics \cite{liu2024smartdampener} and context-aware human-computer interactions \cite{radu2018multimodal}. 

Over the past decade, sensor-based HAR methodologies have profoundly evolved. Early shallow learning paradigms relied on handcrafted statistical and frequency-domain features coupled with traditional classifiers \cite{anguita2013public}. Though effective in controlled settings, they demanded extensive domain expertise and lacked generalization across dynamic human movements. Subsequently, Deep Neural Networks (DNNs) revolutionized the field via automated hierarchical feature extraction. The fusion of Convolutional Neural Networks (CNNs) \cite{zeng2014convolutional} and recurrent architectures---epitomized by DeepConvLSTM \cite{ordonez2016deep}---established robust baselines. Currently, driven by the pursuit of ultra-high precision, the community aggressively embraces large-scale foundation models. Recent breakthroughs exploit heavy Transformer encoders \cite{xu2021limu}, synthesize virtual inertial data via Large Language Models (LLMs) \cite{leng2024imugpt}, and map sensor streams into unified multi-modal embedding spaces \cite{moon2023imu2clip,girdhar2023imagebind}.

However, this relentless pursuit of accuracy and adaptability has birthed a critical bottleneck in wearable computing: the \textbf{arithmetic-latency deadlock}. The current trajectory of DNNs relies heavily on dense 32-bit floating-point (FP32) multiply-accumulate (MAC) operations and complex transcendental functions. In edge hardware, executing these continuous FP32 operations consumes approximately 4.6 pJ per step \cite{horowitz20141}, rapidly draining micro-batteries and rendering always-on HAR highly impractical \cite{ravi2016deep}. Furthermore, these models are trapped in a rigid ``buffer-and-compute'' paradigm, mandating the accumulation of complete temporal windows (e.g., 2 seconds) before inference, fundamentally preventing sub-second responsive interventions. Consequently, there is an urgent necessity to pioneer a paradigm shift towards ``Green HAR''---achieving robust recognition while eradicating extreme power consumption and observation delays \cite{schwartz2020green}.
\begin{figure*}[t]
\centering
 \includegraphics[width=\textwidth]{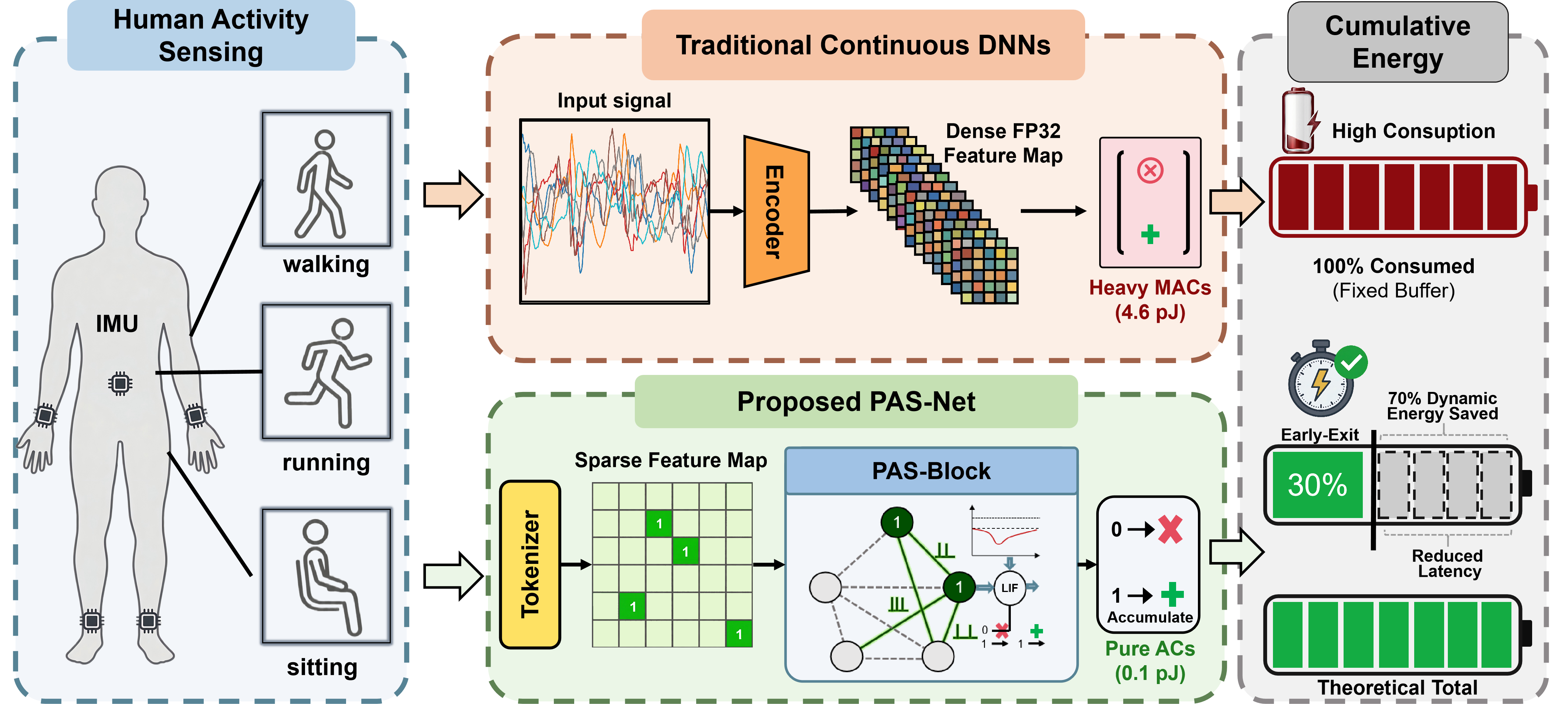}
    % Condense caption to two motivational sentences.
 \caption{\textbf{System overview demonstrating the paradigm shift from traditional, high-latency continuous DNNs to our proposed event-driven PAS-Net.} PAS-Net enables Green Wearable Computing by fundamentally eliminating multipliers in favor of sparse spiking dynamics and unlocking a sub-second streaming early-exit mechanism for continuous IMU streams.}
 \label{fig:paradigm_shift}
\end{figure*}
To shatter this deadlock, Spiking Neural Networks (SNNs) have emerged as a highly promising neuromorphic frontier \cite{maass1997networks}. By processing information through asynchronous binary events (spikes), SNNs replace power-hungry FP32 MACs with multiplier-free accumulate (AC) operations, theoretically reducing energy costs to the magnitude of 0.1 pJ integer additions \cite{zhou2022spikformer, roy2019towards,horowitz20141}. Despite this extreme efficiency, pioneering SNNs for IMU-based HAR remains an uncharted frontier due to severe structural and temporal limitations. First, naive SNNs flatten sensor data, suffering from \textit{spatial blindness} that ignores the biomechanical topology and joint correlations of human movements \cite{wang2023mhagnn}. Second, standard Leaky Integrate-and-Fire (LIF) neurons employ fixed thresholds, lacking the \textit{context-awareness} to adapt to non-stationary movement rhythms \cite{yin2021accurate}. Lastly, training deep SNNs over long IMU streams via Backpropagation Through Time (BPTT) leads to severe gradient degradation, forcing models to rely on long integration windows and failing to resolve the latency crisis.

To bridge these gaps, we propose the \textbf{Physics-Aware Spiking Neural Network (PAS-Net)}, the first fully multiplier-free spiking architecture explicitly tailored for Green HAR (Figure~\ref{fig:paradigm_shift}). PAS-Net fundamentally resolves the aforementioned limitations by decoupling continuous feature extraction from deep spatial-temporal reasoning. It introduces an adaptive symmetric topology to inherently model human biomechanics, alongside an $O(1)$-memory causal neuromodulation mechanism to dynamically adjust neuron firing thresholds. Furthermore, we leverage a temporal-wise training strategy across the continuous spike stream, successfully unlocking an ultra-low latency early-exit mechanism.
The main contributions of this paper are summarized as follows:

\begin{itemize}

    \item \textbf{A Novel Physics-Aware Spiking Paradigm for Green HAR:} We propose PAS-Net, the first multiplier-free spiking architecture explicitly tailored for wearable inertial sensing. By fundamentally shifting the continuous computation paradigm to a purely event-driven mechanism, PAS-Net replaces power-hungry floating-point MACs with sparse, pure AC operations. This effectively shatters the arithmetic bottleneck of traditional DNNs, slashing the operational energy footprint to the magnitude of 0.1 pJ integer additions and establishing a highly viable standard for always-on wearables.

    \item \textbf{Physics- and Context-Aware Adaptive Spiking Dynamics:} To overcome the spatial blindness and neural rigidity of traditional SNNs, we design a dual-adaptive core. Spatially, an Adaptive Symmetric Topology Mixer explicitly enforces human-joint anatomical constraints to suppress small-dataset overfitting. Temporally, we introduce context-aware Dynamic Threshold LIF neurons, modulated by an $O(1)$-memory causal EMA mechanism. This enables neurons to actively adjust their firing sensitivity to non-stationary movement rhythms, mimicking true biological neuromodulation for precise feature extraction.

    \item \textbf{Flexible Streaming Early-Exit via Temporal Supervision:} We break the rigid ``buffer-and-compute'' paradigm inherent to standard block-based ANNs. By innovatively leveraging a Temporal Spike Error (TSE) objective across continuous inertial streams, we unlock a confidence-driven early-exit mechanism. This empowers PAS-Net to achieve state-of-the-art accuracy using only a minimal fraction of the observation window. Consequently, by bypassing redundant computation and full-window buffering, PAS-Net drastically reduces dynamic energy consumption by up to 98\% while ensuring highly responsive wearable interventions.
\end{itemize}

\section{Related Work}
\subsection{Deep Learning Architectures for IMU-based HAR}

The landscape of IMU-based HAR has been continuously refined by architectural evolution. Early benchmarks relied on SVMs coupled with extensive handcrafted features \cite{anguita2013public}. Subsequently, CNNs automated local temporal feature extraction \cite{ronao2016human}, though fusing CNNs with time series features (TSFs) often yielded superior accuracy. For instance, integrating CNNs with 2D FFT \cite{jiang2015human} or statistical features \cite{ignatov2018real} significantly boosted performance.
To capture long-term sequential dynamics, recurrent architectures like LSTM and Bi-LSTM were introduced \cite{li2019bi}. Hybrid models fusing CNN spatial extraction with RNN temporal modeling quickly became the standard, epitomized by DeepConvLSTM \cite{ordonez2016deep}. Further optimizations led to advanced LSTM-CNNs \cite{xia2020lstm} and GRU-based networks (e.g., MchCnnGRU \cite{lu2022multichannel} and ICGNet \cite{dua2023inception}), reaching state-of-the-art performances across complex datasets.
As target activities grew more complex, researchers aggressively integrated Attention mechanisms and advanced macro-architectures to focus on salient spatial-temporal segments. Networks utilizing Inception modules (e.g., iSPLInception \cite{ronald2021isplinception}) or coupling Attention with CNNs/RNNs \cite{khan2021attention,he2018weakly} pushed recognition boundaries. More recently, the community has begun exploring large-scale foundation models (e.g., LIMU-BERT \cite{xu2021limu}) and LLM-driven semantic reasoning \cite{leng2024imugpt,li2025sensorllm} to tackle extreme cross-domain variability and zero-shot scenarios.
However, this relentless scaling of model complexity---peaking with modern foundation models---hits a severe ``power wall.'' High-performing hybrid DNNs, Attention networks, and Transformer encoders rely heavily on dense floating-point matrix multiplications. On resource-constrained wearables, these intensive MAC operations cause unacceptable latency and rapid battery depletion. Developing natively energy-efficient paradigms that maintain strong structural reasoning without massive MAC overhead remains a critical challenge for Green HAR.
\subsection{Energy-Efficient SNNs and Temporal Dynamics}
To circumvent the energy bottlenecks of traditional DNNs, SNNs have emerged as a theoretically superior paradigm for ultra-low-power computing. By processing information via asynchronous, binary events, SNNs fundamentally replace power-hungry MAC operations with highly efficient AC operations \cite{roy2019towards}. Driven by Surrogate Gradient (SG) methods and Spatio-Temporal Backpropagation (STBP) \cite{wu2018spatio}, recent research has transitioned towards deep, complex paradigms. Current advancements in SNN architectures for sequential and time-series modeling broadly follow two trajectories \cite{lv2024efficient}: adapting existing ANN modules to the spiking domain (e.g., Spiking Transformers \cite{zhou2022spikformer, lv2025toward} and Spiking RNNs \cite{bellec2018long}), or designing novel modules governed by biological principles (e.g., Central Pattern Generators \cite{lv2024advancing}).

The fundamental computational unit empowering this temporal processing is typically the Leaky Integrate-and-Fire (LIF) neuron, governed by the following discrete-time dynamics:
\begin{equation}
    U[t] = \tau U[t-1](1 - S[t-1]) + W X[t]
\end{equation}
\begin{equation}
    S[t] = \Theta(U[t] - U_{th})
\end{equation}
where $U[t]$ is the membrane potential at time step $t$, $\tau$ is the leak factor, $X[t]$ is the input current, $W$ represents the synaptic weights, and $S[t]$ is the binary output spike triggered when $U[t]$ exceeds the firing threshold $V_{th}$. $\Theta(\cdot)$ denotes the Heaviside step function. 

While this SG-driven formulation naturally captures temporal dynamics, directly deploying it for highly dynamic IMU-based HAR exposes three critical algorithmic gaps. First, multi-node human movements exhibit intricate spatial correlations \cite{wang2023mhagnn}. Although Spiking Graph Neural Networks (SGNNs) \cite{zhu2022spiking} have been explored for non-Euclidean data, existing SNNs for wearable sensing typically flatten multi-sensor time-series into independent 1D channels, inherently lacking the structural capacity to explicitly route and adapt to dynamic physical topologies. Second, the standard LIF neuron relies on a fixed $V_{th}$. Although adaptive variants (e.g., ALIF \cite{yin2021accurate}) introduce heuristic threshold decays, they still lack causal neuromodulation, struggling to proactively adapt to the fast-changing, causal rhythms of human sports activities. Lastly, despite the success of STBP, optimizing deep SNNs over long time series via Backpropagation Through Time (BPTT) suffers from severe temporal gradient degradation \cite{deng2022temporal}. This forces networks to rely on long integration windows to accumulate gradients, fundamentally precluding ultra-low latency predictions.

Our proposed PAS-Net is explicitly designed to bridge this void. By introducing adaptive physical topology routing and causal neuromodulation, PAS-Net endows SNNs with the spatial-temporal reasoning capabilities of advanced DNNs, while strictly maintaining a multiplier-free, AC-only energy profile.
\section{METHODOLOGY}
\label{sec:methodology}
\begin{figure*}[t]
    \centering
    \includegraphics[width=\textwidth]{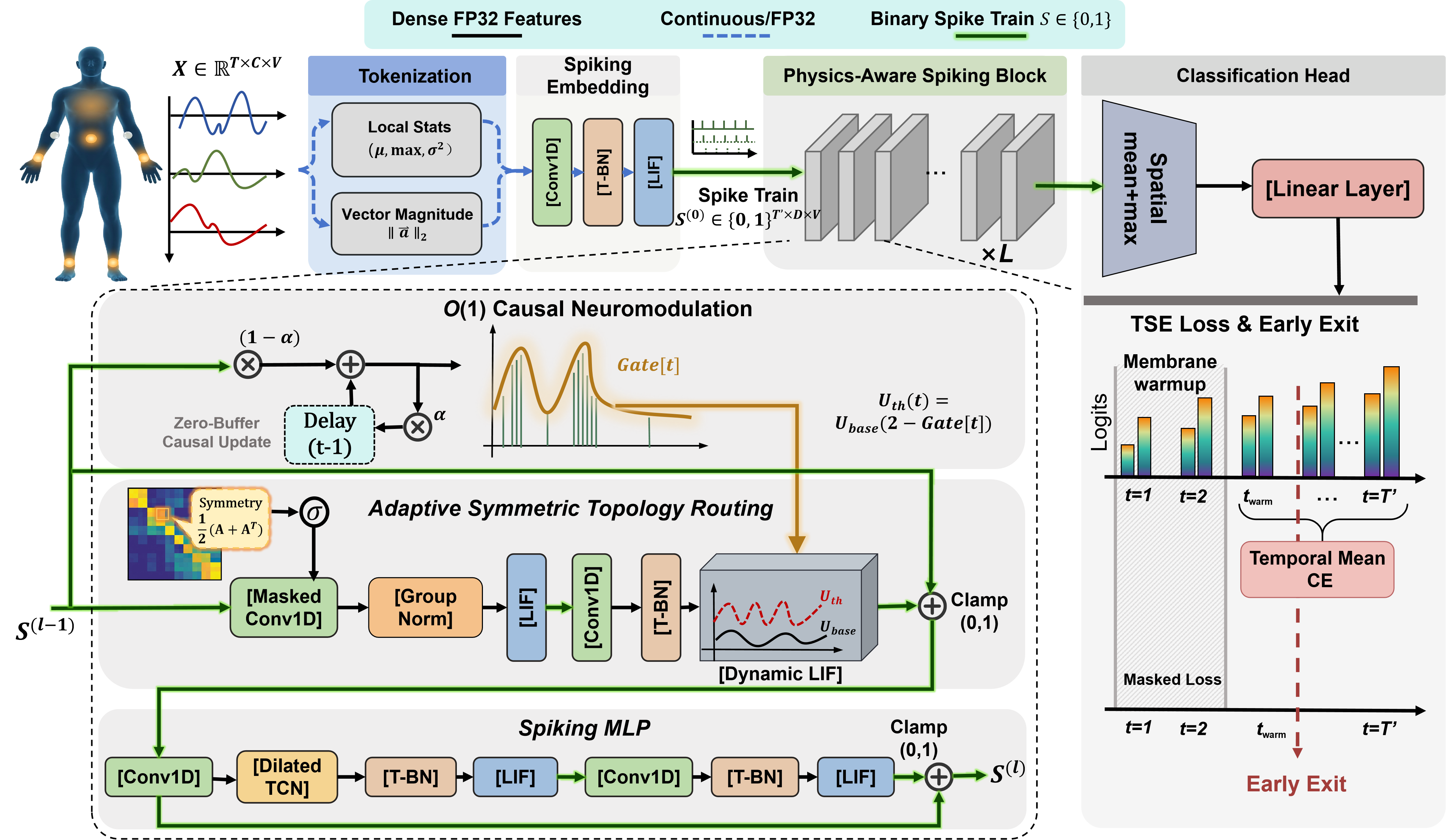}
    \caption{\textbf{Overall architecture of our proposed PAS-Net framework.} A detailed description of PAS-Net is presented in Section \ref{sec:methodology}}
    \label{fig:overall_architecture}
\end{figure*}
\subsection{Overall Architecture of PAS-Net}
To fundamentally address the limitations of existing spiking paradigms for wearable sensing, we propose the Physics-Aware Spiking Network (PAS-Net), a multiplier-free, event-driven architecture designed for Green HAR. Formally, given a raw continuous multi-node IMU tensor $X \in \mathbb{R}^{T \times C \times V}$ (with temporal window $T$, sensor channels $C$, and physical nodes $V$), PAS-Net processes this data through a cohesive, purely spike-driven forward pipeline. 

As illustrated in the system overview (Figure \ref{fig:overall_architecture}), the architecture establishes a clear progression among distinct data domains---seamlessly transitioning from continuous FP32 inputs to binary spike trains ($S \in \{0, 1\}$), causal gating signals, and dense FP32 readout features. Following this left-to-right paradigm, PAS-Net is structured into three cascaded stages. First, in the \textbf{Perceptual Frontend and Spiking Embedding (Sec. \ref{sec:per})}, the raw tensor $X$ is processed by an Invariant Tokenizer to extract rotation-invariant kinematic features and perform structural temporal downsampling. These features are subsequently encoded into a condensed discrete spike train $S^{(0)} \in \{0, 1\}^{T' \times D \times V}$. This pivotal step strategically reduces the sequence horizon and shifts the computation into a sparse, Zero-MAC domain. Next, the binary stream propagates through a deep backbone of $L$ stacked \textbf{Physics-Aware Spiking Blocks (Sec. \ref{sec:PAS})}. These blocks dually adapt to spatial biomechanical linkages via Adaptive Symmetric Topology Routing and to non-stationary temporal rhythms via $O(1)$ Causal Neuromodulation. Finally, in the \textbf{Temporal-Wise Readout and Early-Exit (Sec. \ref{sec:TSE})} stage, a Spatial Mean+Max Pooling head aggregates the features, and a Temporal Spike Error (TSE) objective evaluates the dense FP32 logits at every downsampled time step $t'$. This continuous temporal supervision shatters the traditional ``buffer-and-compute'' latency deadlock, empowering the network to execute ultra-fast early exits without sacrificing accuracy.

\subsection{Perceptual Frontend and Spiking Embedding}
\label{sec:per}
Before extracting deep spatiotemporal correlations, the raw continuous IMU tensor $X \in \mathbb{R}^{T \times C \times V}$ must be robustly aligned, temporally condensed, and encoded into the discrete spike domain. This transformation is achieved through invariant kinematic tokenization and high-dimensional spiking embedding.

\textbf{Invariant Tokenization and Temporal Patching.} Wearable sensors deployed in unconstrained daily scenarios frequently suffer from orientation shifts and device displacement. To mitigate these coordinate-system biases while simultaneously reducing the redundant computational horizon, the Invariant Tokenizer employs a temporal patching strategy. Specifically, at each physical node, the tokenizer computes the rotation-invariant vector magnitude ($L_2$ norm, defined as $\|a\|_2 = \sqrt{a_x^2 + a_y^2 + a_z^2}$) alongside localized statistical features---including the temporal mean, maximum, and variance---extracted over a sliding window with stride $s$. 

Consequently, the original temporal dimension $T$ is mathematically compressed to $T' = \lfloor T / s \rfloor$. This projection enriches the raw input into a robust, dense continuous token tensor $X_{tok} \in \mathbb{R}^{T' \times C_{tok} \times V}$, where $C_{tok}$ denotes the expanded channel dimension of the concatenated statistical features. By explicitly downsampling the sequence, this mechanism not only expands the causal receptive field for downstream spiking neurons but also exponentially cuts down the total Synaptic Operations (SOPs), establishing the foundation for our ultra-low-power footprint.

\textbf{Temporal-Aware Normalization and Spiking Embedding.} To initialize the multiplier-free execution pipeline, the continuous condensed tensor $X_{tok}$ must be projected into a high-dimensional latent space and discretized. The Spiking Embedding block accomplishes this via a cascaded 1D temporal convolution, a Temporal-Aware Batch Normalization (T-BN) layer, and Leaky Integrate-and-Fire (LIF) neurons.

A critical component bridging the continuous and spiking domains is the T-BN layer. Naively applying standard frame-wise Batch Normalization independently at each time step artificially rescales inactive frames and suppresses motion bursts, fundamentally destroying the relative temporal rhythm of human kinematics. To resolve this, T-BN computes the global statistics by pooling across both the condensed temporal window $T'$ and the spatial nodes $V$:
\begin{equation}
\tilde{X}_{tok}[t'] = \gamma \frac{X_{tok}[t'] - \mu_{global}}{\sqrt{\sigma_{global}^2 + \epsilon}} + \beta    
\end{equation}
where $X_{tok}[t']$ and $\tilde{X}_{tok}[t']$ denote the linearly projected features and the normalized features at the downsampled time step $t' \in [1, T']$, respectively. The terms $\mu_{global}$ and $\sigma_{global}^2$ represent the mean and variance computed globally across the $T'$ and $V$ dimensions. The variables $\gamma$ and $\beta$ are learnable affine parameters, and $\epsilon$ is a stability constant. By sharing these global spatiotemporal statistics, T-BN stabilizes the membrane potential updates while strictly preserving the peak-to-valley energy ratio of the motion sequences. 

Finally, the normalized continuous features $\tilde{X}_{tok}[t']$ are integrated into the membrane potential of the initial LIF neurons. Upon exceeding a predefined voltage threshold, these neurons emit binary events, officially converting the features into the initial spike train tensor $S^{(0)} \in \{0, 1\}^{T' \times D \times V}$, where $D$ is the specified embedding dimension. This transformation marks the strict boundary where dense floating-point operations terminate.

\subsection{Physics-Aware Spiking Block (PAS-Block)}
\label{sec:PAS}
The core computational engine of our network relies on $L$ stacked Physics-Aware Spiking Blocks (PAS-Blocks). Inspired by modern macro-architectures, each PAS-Block is designed as a two-stage sequential pipeline: a spatial topology token mixer followed by a temporal channel mixer. 

Let $S^{(l-1)} \in \{0, 1\}^{T' \times D \times V}$ denote the input binary spike tensor to the $l$-th PAS-Block, where $l \in \{1, 2, \dots, L\}$. To maintain mathematical brevity, the following equations omit spatial and channel subscripts, as all operations are natively executed as element-wise or broadcasted tensor computations.

\subsubsection{Stage 1: Physics Topology Routing and Dynamic Modulation}
The first stage focuses on capturing the spatial biomechanical linkages between the $V$ physical nodes while proactively adapting to non-stationary movement rhythms. Upon receiving $S^{(l-1)}$, the data flow diverges into two parallel branches: causal neuromodulation and structural topology routing, which subsequently converge at the dynamic spiking activation.

\textbf{O(1) Causal Neuromodulation.} Human sports activities typically alternate between relative stasis and explosive bursts. To capture this energy envelope without incurring the massive memory overhead of sliding-window buffers, a parallel causal Exponential Moving Average (EMA) neuromodulator extracts the dynamic rhythm directly from the input spikes. At each time step $t$, it updates a causal energy gate $Gate^{(l)}[t] \in [0, 1]$ using a single-step delayed hidden state:
\begin{equation}
    Gate^{(l)}[t] = \alpha \cdot Gate^{(l)}[t-1] + (1 - \alpha) \cdot S^{(l-1)}[t]
\end{equation}
where $\alpha \in (0, 1)$ is a learnable decay factor. This zero-buffer extraction maintains strict causality and extreme lightweight efficiency for real-time streaming.

\textbf{Adaptive Symmetric Topology Routing.} Concurrently, the network models the $V$ distinct IMU nodes as a dynamic physical graph to exchange spatial kinematics. To explicitly respect the bidirectional physical nature of human joints and prevent overfitting to asymmetric shortcuts on small datasets, a learnable adjacency matrix $\mathbf{A}^{(l)}$ is constrained to be strictly symmetric:
$\tilde{\mathbf{A}}^{(l)} = \sigma \left( \frac{\mathbf{A}^{(l)} + (\mathbf{A}^{(l)})^\top}{2} \right)$
where $\sigma(\cdot)$ denotes the Sigmoid activation function, and $\tilde{\mathbf{A}}^{(l)} \in \mathbb{R}^{V \times V}$ is the effective symmetric adjacency mask. 

Instead of isolating spatial routing and temporal integration into separate, computationally expensive steps, we design a highly efficient \textit{Masked Spatiotemporal Convolution}. Specifically, the topology mixer is formulated as a 1D causal convolution applied along the temporal axis, which is shared across all hidden feature channels to minimize parameter overhead. The spatial mask $\tilde{\mathbf{A}}^{(l)}$ is broadcasted and element-wise multiplied with the temporal kernel weights $\mathbf{W}_{topo}^{(l)} \in \mathbb{R}^{V \times V \times k}$:
\begin{equation}
    I_{topo}^{(l)}[t] = \sum_{j=0}^{k-1} \left( \mathbf{W}_{topo}^{(l)}[j] \odot \tilde{\mathbf{A}}^{(l)} \right) S^{(l-1)}[t - j]
\end{equation}
where $k$ is the temporal kernel size (empirically set to $k=5$) with strictly causal padding. This unified operation simultaneously routes spatial physical dependencies and integrates short-term micro-motion trajectories over the temporal window. Crucially, because the incoming tensor $S^{(l-1)} \in \{0, 1\}$, this masked spatiotemporal convolution mathematically reduces to sparse Accumulate (AC-only) operations, explicitly eliminating power-hungry floating-point MAC units.

\textbf{True Dynamic LIF and Intermediate Residual.} The parallel branches now converge. The topologically routed current is first projected via a 1D convolution and T-BN to form the structured input current $I_{dyn}^{(l)}[t]$. This current feeds into our True Dynamic LIF neurons to generate the intermediate spatial spike tensor $S_{topo}^{(l)}[t]$. Under the active control of the neuromodulation gate, the temporal dynamics of the membrane potential $U^{(l)}[t]$ and the dynamic threshold $U_{th}^{(l)}[t]$ are formulated as:
\begin{equation} \label{eq:membrane_update}
    U^{(l)}[t] = \tau U^{(l)}[t-1] \odot \left(1 - S_{topo}^{(l)}[t-1]\right) + I_{dyn}^{(l)}[t]
\end{equation}
\begin{equation} \label{eq:threshold_and_spike}
   U_{th}^{(l)}[t] = U_{base} \left( 2 - Gate^{(l)}[t] \right), \quad S_{topo}^{(l)}[t] = \Theta \left(U^{(l)}[t] - U_{th}^{(l)}[t] \right)
\end{equation}
where $\tau$ is the leak constant, $U_{base}$ is the static baseline threshold, and $\Theta(\cdot)$ is the Heaviside step function. Note that $S_{topo}^{(l)}[t-1]$ in Eq.~\ref{eq:membrane_update} acts as a recurrent self-reset mechanism. The threshold formulation in Eq.~\ref{eq:threshold_and_spike} elegantly maps the $[0, 1]$ energy envelope into an inverted threshold multiplier, scaling between $U_{base}$ (highest sensitivity for burst motions) and $2U_{base}$ (noise suppression for idle states).

To preserve the pure-spike data format, a bounding residual connection generates the mid-level feature tensor:
\begin{equation}
    S_{mid}^{(l)}[t] = \text{Clamp} \left( S^{(l-1)}[t] + S_{topo}^{(l)}[t], \ 0, \ 1 \right)
\end{equation}
where the $\text{Clamp}(\cdot, 0, 1)$ function serves as a logical OR-gate, preventing activation accumulation from breaking the binary data flow.

\subsubsection{Stage 2: Spiking Dilated Temporal Convolutions (TCN)}
The second stage acts as a temporal and channel mixer. The mid-level spike tensor $S_{mid}^{(l)}$ enters the Spiking MLP branch to construct deep temporal receptive fields along the time axis $T$.

To efficiently aggregate long-term historical context without violating causal constraints, this branch employs Spiking Dilated TCNs. The discrete causal temporal convolution with an exponentially increasing dilation factor $d$ is mathematically defined to compute the temporal current $I_{mlp}^{(l)}[t]$:
\begin{equation}
    I_{mlp}^{(l)}[t] = \sum_{k=0}^{K-1} \mathbf{W}_{tcn}^{(l)}[k] S_{mid}^{(l)}[t - k \cdot d]
\end{equation}
where $\mathbf{W}_{tcn}^{(l)}$ denotes the learnable 1D convolution kernel of size $K$, and $d$ is the layer-specific dilation rate (e.g., $d = 2^{l-1}$). By skipping intermediate steps, the network expands its temporal horizon while maintaining linear computational scaling.

This temporal current then drives the subsequent MLP-branch LIF neurons to generate the temporal spike tensor $S_{mlp}^{(l)}[t]$:
\begin{equation}
    U_{mlp}^{(l)}[t] = \tau U_{mlp}^{(l)}[t-1] \odot \left(1 - S_{mlp}^{(l)}[t-1]\right) + I_{mlp}^{(l)}[t]
\end{equation}
\begin{equation}
    S_{mlp}^{(l)}[t] = \Theta\left(U_{mlp}^{(l)}[t] - U_{base}\right)
\end{equation}
where $U_{mlp}^{(l)}[t]$ is the membrane potential of the temporal mixer. 

Finally, a second bounding residual connection integrates these temporal features to produce the ultimate output tensor of the $l$-th PAS-Block, $S^{(l)}[t]$:
\begin{equation}
    S^{(l)}[t] = \text{Clamp} \left( S_{mid}^{(l)}[t] + S_{mlp}^{(l)}[t], \ 0, \ 1 \right)
\end{equation}
This mathematically guarantees that the output tensor $S^{(l)}$ remains strictly binary $\{0, 1\}$, seamlessly serving as the Zero-MAC input for the subsequent layer and securing an unbroken spike-driven paradigm.
\subsection{Temporal-Wise Readout and Ultra-Low Latency Strategy}
\label{sec:TSE}
\subsubsection{Spatial Mean+Max Pooling}
Upon exiting the stacked PAS-Blocks, the feature map must be aggregated across the $V$ physical nodes. Standard mean pooling tends to dilute the discriminative kinematic signals of highly active joints (e.g., the wrist during a tennis serve) with the noise of static joints (e.g., the non-dominant hand). Conversely, pure max pooling may discard valuable contextual movements. PAS-Net employs a fused Mean+Max Spatial Pool ($X_{pool} = Mean(X, dim=V) + Max(X, dim=V)$), maintaining the continuous time dimension $T$ while optimally preserving prominent motion features.

\subsubsection{Temporal Spike Error (TSE) Loss with Membrane Warmup}
Standard sequence modeling paradigms accumulate features over the entire temporal window $T$ to generate a single prediction at the final step, inherently causing high decision delays and severe gradient degradation via Backpropagation Through Time (BPTT).

PAS-Net leverages a Temporal Spike Error (TSE) Loss framework. The network projects the pooled spike representations into a continuous sequence of logits, generating a valid classification prediction $\hat{y}[t]$ at every time step $t$. To prevent the initial unstable membrane potentials (prior to sufficient current accumulation) from polluting the gradients, TSELoss introduces a membrane warmup mechanism, masking the loss for the initial $t_{warm}$ steps:
\begin{equation}
    \mathcal{L}_{TSE} = \frac{1}{\sum_{t=t_{warm}}^{T} w_t} \sum_{t=t_{warm}}^{T} w_t \mathcal{L}_{CE}(\hat{y}[t], y)
\end{equation}
where $\mathcal{L}_{CE}(\cdot)$ denotes the standard Cross-Entropy loss, $y$ represents the ground-truth label, and $\hat{y}[t]$ is the predicted class probability distribution at time step $t$. The term $t_{warm}$ specifies the predefined number of initial warmup steps, and $w_t$ is a time-varying weighting factor (e.g., linear escalation) applied at step $t$. 

This temporal-wise supervision provides dense, short-path gradients that stabilize the optimization of deep spiking dynamics. More profoundly, it continuously penalizes misclassifications at early stages, empowering the edge device to make highly confident predictions using only a fraction of the observation window. Consequently, PAS-Net shatters the traditional ``buffer-and-compute'' deadlock, enabling sub-second, ultra-low latency early exits without sacrificing accuracy.
\section{EVALUATION}
\label{sec:evaluation}

To comprehensively evaluate the performance and system-level efficiency of the proposed PAS-Net, we conduct extensive experiments across multiple established benchmarks. Our evaluation aims to answer the following research questions (RQs):
\begin{itemize}
    \item \textbf{RQ1 (Recognition Accuracy):} How does PAS-Net perform compared to state-of-the-art DNNs and SNNs in subject-independent HAR scenarios?
    \item \textbf{RQ2 (Energy Efficiency):} To what extent does the event-driven spiking dynamic reduce the theoretical and hardware-level energy consumption?
    \item \textbf{RQ3 (Latency Reduction):} Can the temporal-wise supervision and early-exit mechanism effectively reduce the observation latency without compromising accuracy?
\end{itemize}

\subsection{Experimental Setup}

\subsubsection{Datasets}
To ensure the robustness of our evaluation across diverse wearable scenarios, we select seven publicly available IMU datasets. As summarized in Table \ref{tab:datasets}, these datasets cover a wide spectrum of physical sensor topologies, sampling rates, and target demographics:
\begin{itemize}
    \item \textbf{Multi-node Sensor Environments:} \textit{PAMAP2} \cite{reiss2012introducing}, \textit{Daily-Sports} \cite{altun2010comparative}, and \textit{TNDA-HAR}\cite{4epb-pg26-21} feature complex daily and sports activities captured by multiple distributed sensors. Notably, \textit{TNDA-HAR} incorporates a massive cohort of 50 subjects wearing 6 asymmetrical sensor nodes across the torso and limbs. Furthermore, \textit{HuGaDB} \cite{chereshnev2017hugadb} provides detailed continuous gait recordings using a dense network of 6 inertial sensors on the lower body. Together, these datasets provide a rigorous, highly-coupled spatial testbed for evaluating our adaptive physical topology routing.
    \item \textbf{Single-node Daily Locomotion:} \textit{USC-HAD} \cite{zhang2012usc} serves as a standard foundational benchmark for evaluating basic human locomotion (e.g., walking, jumping) using a single front-right hip sensor.
    \item \textbf{Clinical and Older Adult Demographics:} To validate the context-awareness of our model on non-stationary and irregular movements, we include the \textit{Daphnet Parkinson's Freezing of Gait (FOG)} dataset \cite{bachlin2009wearable} and the \textit{HAR70+} dataset \cite{ustad2023validation}, which capture the unique kinematic rhythms of Parkinson's patients and older adults (aged 70+), respectively.
\end{itemize}

% --- Dataset Summary Table ---
\begin{table*}[htbp]
\centering
\caption{Summary of the selected IMU datasets. Window sizes ($T$) and strides are adaptively configured based on the native sampling rates to maintain a consistent physical observation duration (approx. 2 seconds), with the exception of Daily-Sports, which uses full 5-second episodic segments.}
\label{tab:datasets}
\resizebox{\textwidth}{!}{
\begin{tabular}{lccccccc}
\toprule
\textbf{Dataset} & \textbf{Subjects} & \textbf{Classes} & \textbf{Sensor Nodes ($V$)} & \textbf{Sampling Rate} & \textbf{Window Size ($T$)} & \textbf{Stride} & \textbf{Physical Duration} \\
\midrule
\textbf{TNDA-HAR}\cite{4epb-pg26-21}  & 50 & 8 & 6 & $\sim$ 100 Hz & 200 & 100 & $\sim$ 2.0 s \\
\textbf{PAMAP2}\cite{reiss2012introducing} & 9 & 18 & 3 & 100 Hz & 200 & 100 & $\sim$ 2.0 s \\
\textbf{Daily-Sports}\cite{altun2010comparative} & 8 & 19 & 5 & 25 Hz & 125 & N/A & 5.0 s (Full Segment) \\
\textbf{HuGaDB}\cite{chereshnev2017hugadb} & 18 & 12 & 6 & $\sim$ 50 Hz & 100 & 50 & $\sim$ 2.0 s \\
\textbf{USC-HAD}\cite{zhang2012usc} & 14 & 12 & 1 & 100 Hz & 200 & 100 & $\sim$ 2.0 s \\
\textbf{HAR70+}\cite{ustad2023validation} & 18 & 7 & 2 & 50 Hz & 100 & 50 & $\sim$ 2.0 s \\
\textbf{Parkinson (FOG)}\cite{bachlin2009wearable} & 10 & 2 & 1 & $\sim$ 64 Hz & 128 & 64 & $\sim$ 2.0 s \\
\bottomrule
\end{tabular}
}
\end{table*}

\subsubsection{Data Preprocessing and Adaptive Windowing}
In practical wearable applications, identical physical activities manifest differently across hardware platforms due to varying sampling rates. To prevent frequency distortion, we avoid enforcing a uniform algorithmic window size. Instead, we implement a \textit{frequency-adaptive windowing strategy}. As detailed in Table \ref{tab:datasets}, the sliding window size $T$ and stride are dynamically adjusted to the native sampling rate of each dataset, ensuring a consistent physical observation window of approximately 2 seconds with a 50\% overlap. For the \textit{Daily-Sports} dataset, we respect its original episodic protocol by processing the pre-segmented 5-second files directly without additional sliding windows.

\subsubsection{Evaluation Protocol: Subject-Independent Split}
A common pitfall in HAR evaluation is the use of window-level random splitting, which inadvertently mixes temporal segments from the same subject into both training and testing sets, leading to severe data leakage and inflated performance. To rigorously assess the generalization capability of PAS-Net on unseen users---a critical requirement for real-world deployments---we adopt a strict \textbf{Subject-Independent Hold-out Split}.

For each dataset, the data is partitioned entirely based on unique Subject IDs (or physical session identifiers). We allocate 70\% of the subjects for training, 15\% for validation, and 15\% for testing. Under this protocol, the test set consists exclusively of kinematic patterns from individuals unseen during the training phase. All results are reported using the Macro F1-score to account for class imbalances commonly found in clinical and daily datasets.

\subsubsection{Hardware Energy Estimation Protocol}
To quantify the system-level energy efficiency beyond theoretical FLOP counts, we map the algorithmic operations to established 45nm CMOS hardware logic measurements \cite{horowitz20141, panda2020toward}. For baseline continuous DNNs, standard multiply-accumulate (MAC) operations require 32-bit floating-point (FP32) precision to prevent overflow. A single FP32 MAC operation consumes $E_{MAC} = 4.6$ pJ (combining $3.7$ pJ for multiplication and $0.9$ pJ for addition). Consequently, the total inference energy for a conventional DNN is estimated as:
\begin{equation}
    E_{DNN} = FLOPs_{total} \times E_{MAC}
\end{equation}

In contrast, PAS-Net restricts continuous signals strictly to the shallow input tokenizer. Its deep spatial-temporal core operates purely on binary spikes ($S \in \{0, 1\}$), which fundamentally bypasses power-hungry multiplier circuits and reduces synaptic operations to 32-bit integer accumulations (AC). Each spike-triggered AC operation consumes only $E_{AC} = 0.1$ pJ. Therefore, the total inference energy of PAS-Net is dynamically determined by the actual event-driven sparsity, calculated as:
\begin{equation}
    E_{PAS\text{-}Net} = FLOPs_{stem} \times E_{MAC} + SOPs_{core} \times E_{AC}
\end{equation}
where $FLOPs_{stem}$ denotes the continuous operations bounded within the first encoding layer, and $SOPs_{core}$ represents the total Synaptic Operations within the deep spiking blocks. Specifically, the SOPs are intrinsically tied to the layer-wise spike firing rate $Fr$: $SOPs_{core} = \sum_{l=1}^{L} (FLOPs_l \times Fr_l)$. By intelligently suppressing redundant spike emissions ($Fr \ll 1$) via causal neuromodulation, PAS-Net translates its mathematical sparsity directly into quantifiable, ultra-low hardware energy metrics.
\subsubsection{Implementation Details}
All baseline models and PAS-Net are implemented in PyTorch using the SpikingJelly framework and trained from scratch on four NVIDIA V100 GPUs under a unified protocol. We optimize using AdamW (learning rate: $2.8 \times 10^{-4}$ to $1 \times 10^{-3}$, weight decay: $\le 0.01$) alongside a Cosine Annealing scheduler with a 12--20 epoch linear warmup, decaying to a minimum of $1 \times 10^{-7}$. Models are trained for up to 220 epochs with a batch size of 32--48. To mitigate overfitting, label smoothing (0.1) and Mixup ($\alpha \in [0.2, 0.5]$) are uniformly applied. Model selection is strictly determined by peak validation accuracy before final independent testing. Notably, the Temporal Spike Error (TSE) loss and early-exit inference are exclusively applied to PAS-Net. To prevent early noisy gradients from perturbing deep spiking dynamics, TSE utilizes a dynamic weighting strategy with a membrane warmup ratio ($t_{warm}$) spanning 15\%--25\% of the total temporal window $T$.
\subsection{Overall Recognition Performance}

To address RQ1, we comprehensively evaluate the recognition accuracy of the proposed PAS-Net against a vast array of baseline models, including 7 state-of-the-art continuous DNNs and 11 event-driven SNNs. Table \ref{tab:overall_accuracy} presents the detailed performance across seven public HAR datasets under the strict subject-independent hold-out protocol. 

\textbf{Superiority among Event-Driven SNNs.} 
The results clearly establish PAS-Net as the new state-of-the-art within the neuromorphic computing domain for wearable sensing. Across all evaluated benchmarks, PAS-Net consistently and significantly outperforms existing spiking architectures, ranging from basic multi-layer perceptrons (SNN MLP) to advanced attention-based models (e.g., STAtten, Spikformer, and QKFormer). Notably, on highly complex multi-node datasets such as PAMAP2 and TNDA-HAR, PAS-Net achieves absolute accuracy improvements of $\sim 12\%$ and $\sim 7\%$ over the strongest SNN baselines, respectively. Furthermore, on the foundational USC-HAD dataset, our model records a massive $16\%$ performance leap over the Spikformer architecture. These substantial gains validate the effectiveness of our Adaptive Symmetric Topology Routing, which explicitly models the biomechanical dependencies of human joints rather than treating spatial nodes as isolated features.

\textbf{Closing the Gap with Continuous DNNs.} 
A long-standing stereotype in neuromorphic computing is the inevitable trade-off between energy efficiency and recognition accuracy, with SNNs traditionally lagging behind heavy continuous DNNs. PAS-Net fundamentally shatters this barrier. As shown in Table \ref{tab:overall_accuracy}, PAS-Net not only matches but frequently surpasses highly parameterized, floating-point DNN baselines. For instance, on PAMAP2 and USC-HAD, PAS-Net outperforms the best-performing DNNs (ResNet-SE and DeepConvLSTM) by $+7.24\%$ and $+8.99\%$, respectively. On TNDA-HAR, our event-driven architecture outperforms the heaviest self-attention networks (Self-HAR). 

We do observe that on the Daily-Sports dataset, heavily parameterized ANNs (e.g., Self-HAR at $88.73\%$) retain an edge over PAS-Net ($82.06\%$). This is primarily due to the episodic, pre-segmented 5-second nature of this dataset, which perfectly aligns with the global temporal aggregation strengths of deep Transformers. Nevertheless, PAS-Net still dominates all other SNNs on this dataset by a wide margin (outperforming the second-best SNN, SpikeDrivenTransformerV2, by over $3\%$). In clinical scenarios such as the Parkinson (FOG) dataset, both mainstream DNNs and PAS-Net achieved perfect recognition ($100\%$), demonstrating that the spiking paradigm can flawlessly capture distinct pathological rhythms without precision loss.
Note that the performance of certain benchmarked DNNs, such as rTsfNet, appears lower than the results reported in their original literature \cite{lv2024efficient}. This is a direct consequence of our stringent evaluation protocol: 1) we employ a strict Subject-Independent split, which eliminates temporal data leakage common in window-level shuffling; and 2) to ensure a fair end-to-end comparison, we feed raw sensor signals to all models, whereas rTsfNet originally relies on extensive manual feature engineering (45 discrete TSFs) \cite{lv2024efficient}. These results further underscore the robustness of PAS-Net, demonstrating its ability to maintain high precision even in highly challenging, feature-deprived, and cross-subject scenarios.

% --- Overall Accuracy Table ---
\begin{table*}[htbp]
\centering
\caption{Overall recognition performance (Accuracy \%) on seven diverse IMU datasets under the subject-independent evaluation protocol. The best results among all models are highlighted in \textbf{bold}, and the second best are \underline{underlined}.}
\label{tab:overall_accuracy}
\resizebox{\textwidth}{!}{
\begin{tabular}{l|l|ccccccc}
\toprule
\textbf{Paradigm} & \textbf{Model Architecture} & \textbf{PAMAP2} & \textbf{Daily-Sports} & \textbf{TNDA} & \textbf{HuGaDB} & \textbf{USC-HAD} & \textbf{HAR70+} & \textbf{Parkinson} \\
\midrule
\multirow{7}{*}{\textbf{\shortstack{Continuous \\ DNNs}}} 
& DeepConvLSTM \cite{ordonez2016deep} & 70.23 & 84.31 & 92.16 & \underline{92.80} & \underline{64.68} & \underline{88.31} & \textbf{100.00} \\
& ResNet-SE\cite{hu2018squeeze,he2016deep}& \underline{76.31} & \underline{86.86} & 92.60 & 92.48 & 64.06 & \textbf{88.36} & \textbf{100.00} \\
& MCHCNN-GRU\cite{lu2022multichannel}& 75.35 & 83.82 & 91.69 & \textbf{93.40} & 63.73 & 87.63 & 99.97 \\
& Self-HAR\cite{tang2021selfhar} & 71.55 & \textbf{88.73} & \underline{93.00} & 92.77 & 58.89 & 88.09 & \textbf{100.00} \\
& Uni-HAR\cite{xu2023practically} & 46.22 & 53.53 & 36.37 & 72.37 & 30.40 & 78.20 & \textbf{100.00} \\
& rTsfNet\cite{enokibori2024rtsfnet} & 60.98 & 48.63 & 74.44 & 77.91 & 40.82 & 77.70 & \textbf{100.00} \\
& IFConvTransformer\cite{zhang2022if}& 67.11 & 74.41 & 88.29 & 91.74 & 58.58 & 85.47 & \textbf{100.00} \\
\midrule
\multirow{12}{*}{\textbf{\shortstack{Event-Driven \\ SNNs}}} 
& SNN MLP & 66.07 & 69.71 & 74.20 & 89.20 & 44.29 & 76.71 & \textbf{100.00} \\
& LMU-Former\cite{liu2024lmuformer} & 44.78 & 63.53 & 73.09 & 90.16 & 54.47 & 77.78 & \textbf{100.00} \\
& QKFormerIMU\cite{zhou2024qkformer} & 66.43 & 76.76 & 83.50 & 90.19 & 56.18 & 82.79 & \textbf{100.00} \\
& STAtten\cite{lee2025spiking} & 70.79 & 72.06 & 87.69 & 90.72 & 55.30 & 84.48 & \textbf{100.00} \\
& SpikeDrivenTransformer\cite{yao2023spike}& 70.59 & 73.92 & 86.15 & 91.46 & 54.29 & 84.16 & \textbf{100.00} \\
& SpikeDrivenTransformerV2\cite{yao2024spike} & 68.99 & 77.35 & 87.53 & 91.85 & 54.66 & 84.62 & \textbf{100.00} \\
& SpikeGRU\cite{lv2024efficient}& 13.65 & 6.08 & 14.60 & 58.26 & 22.43 & 40.99 & \textbf{100.00} \\
& SpikeRNN\cite{lv2024efficient}& 67.71 & 68.63 & 81.72 & 92.10 & 40.31 & 84.98 & \textbf{100.00} \\
& SpikeTCN\cite{lv2024efficient}& 64.19 & 75.69 & 90.58 & 90.83 & 55.05 & 83.45 & \textbf{100.00} \\
& Spikformer\cite{zhou2022spikformer}& 64.59 & 73.92 & 80.93 & 89.54 & 57.62 & 83.90 & \textbf{100.00} \\
& TSSNN\cite{lv2024efficient}& 70.63 & 64.71 & 79.90 & 84.36 & 39.92 & 78.85 & \textbf{100.00} \\
\rowcolor{gray!10}
& \textbf{PAS-Net (Ours)} & \textbf{83.55} & 82.06 & \textbf{95.25} & \underline{92.80} & \textbf{73.67} & 87.45 & \textbf{100.00} \\
\bottomrule
\end{tabular}
}
\end{table*}
\begin{figure*}[htbp]
    \centering
    
    % ================= 1. PAMAP2: DNN =================
    \begin{subfigure}[b]{0.24\textwidth}
        \centering
        \includegraphics[width=\linewidth]{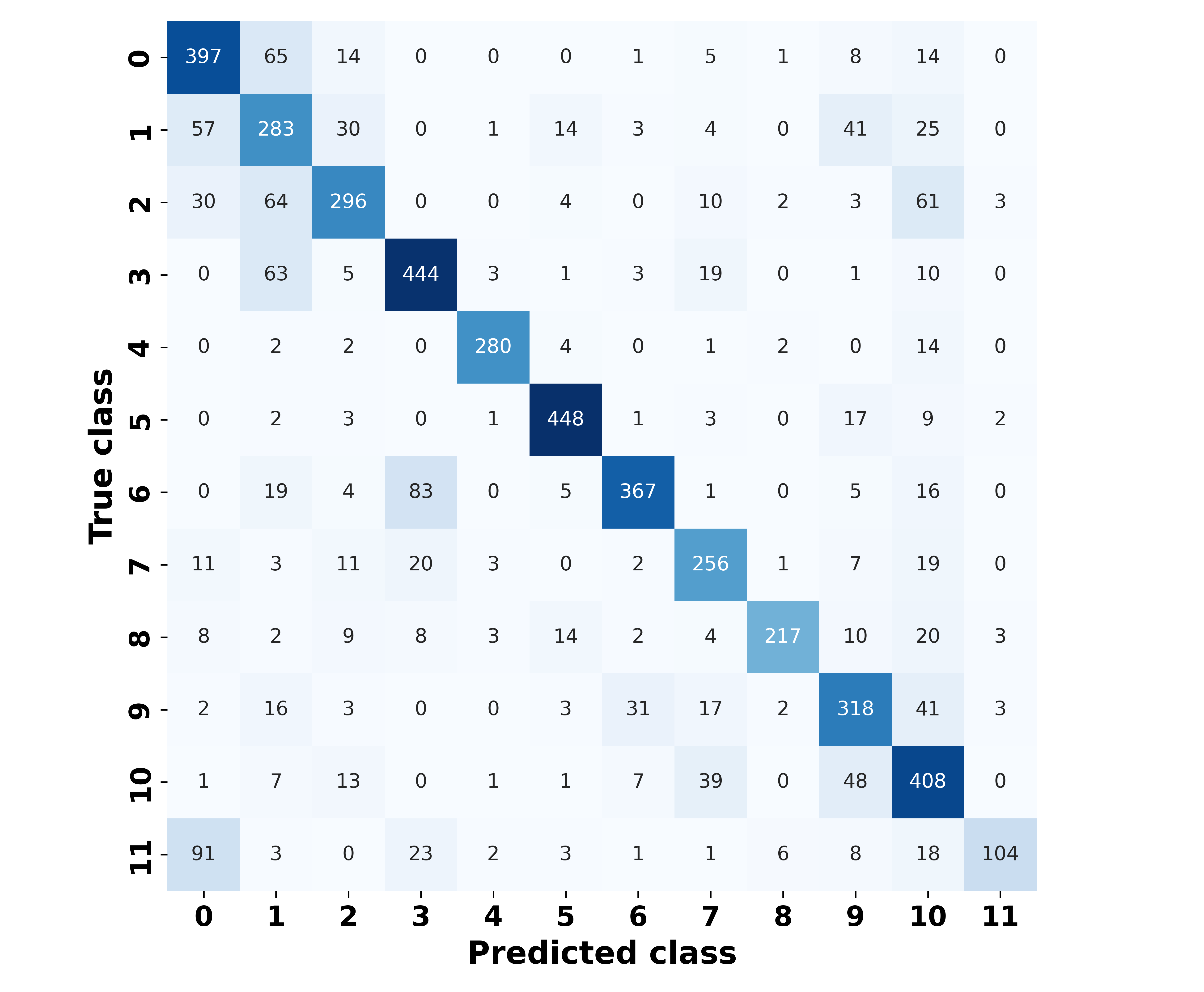}
        \caption{PAMAP2: DeepConvLSTM}
        \label{fig:cm_pamap2_dnn}
    \end{subfigure}
    \hfill
    % ================= 2. PAMAP2: PAS-Net =================
    \begin{subfigure}[b]{0.24\textwidth}
        \centering
        \includegraphics[width=\linewidth]{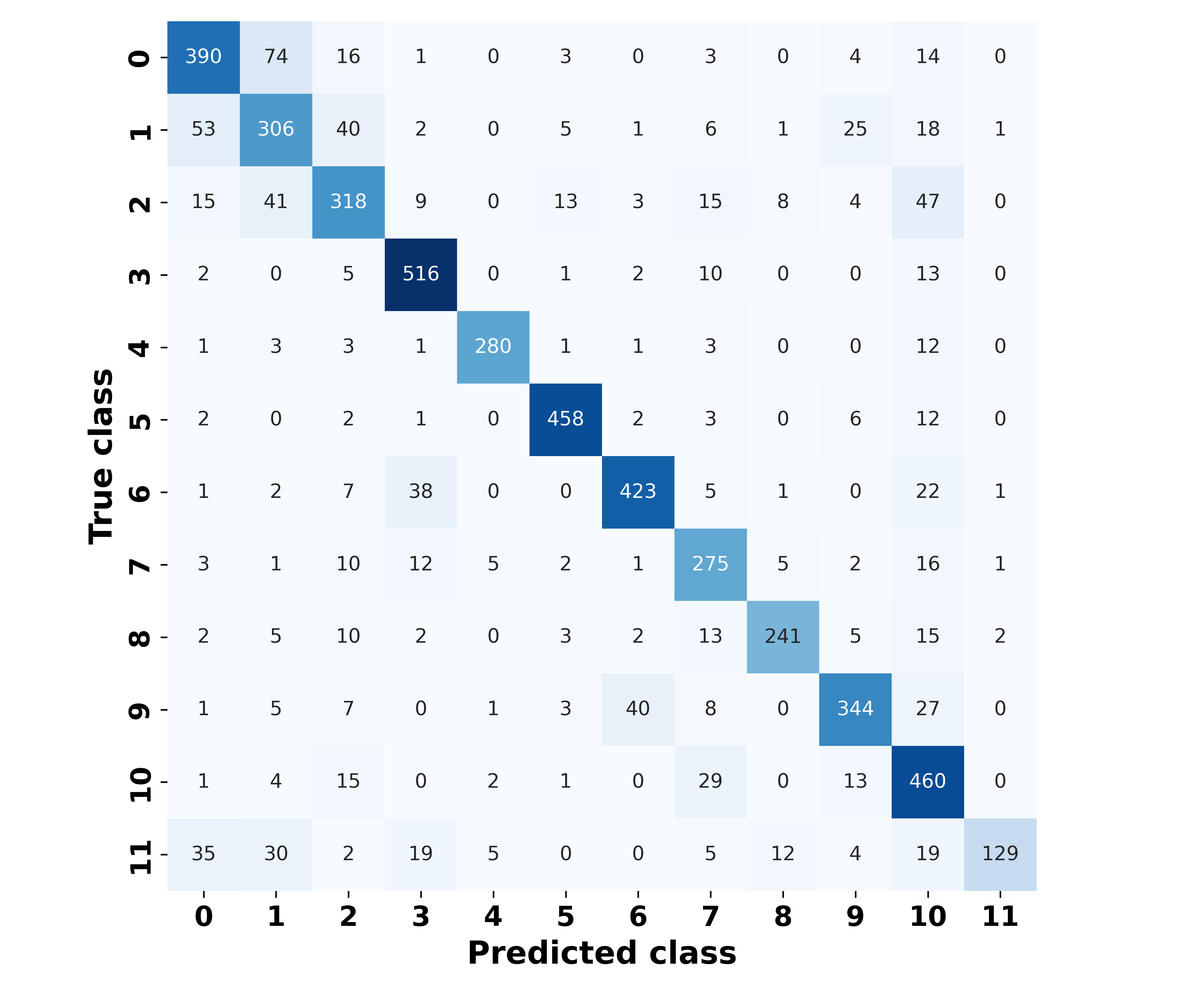}
        \caption{PAMAP2: PAS-Net (Ours)}
        \label{fig:cm_pamap2_pasnet}
    \end{subfigure}
    \hfill
    % ================= 3. TNDA: DNN =================
    \begin{subfigure}[b]{0.24\textwidth}
        \centering
        \includegraphics[width=\linewidth]{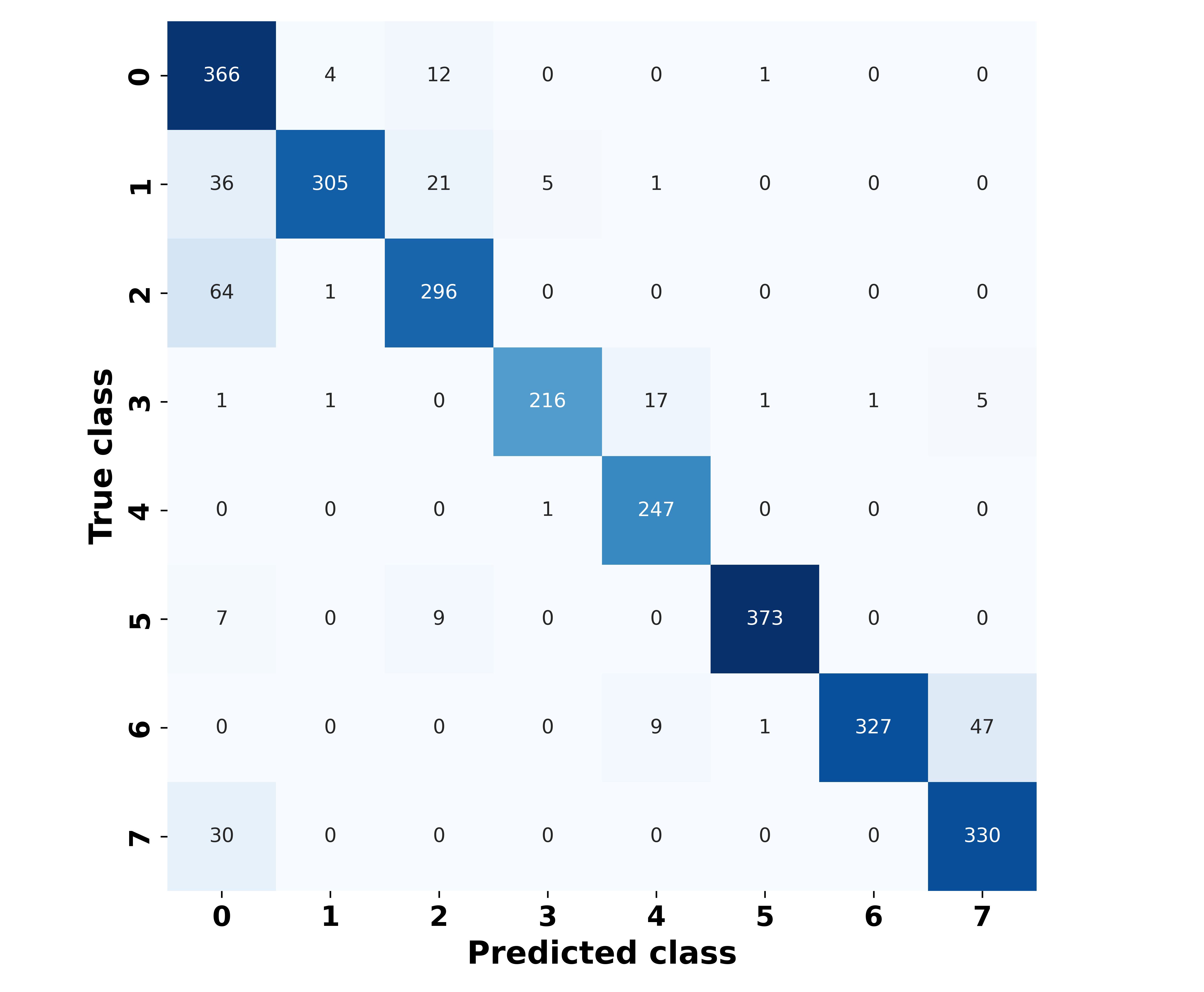}
        \caption{TNDA: DeepConvLSTM}
        \label{fig:cm_tnda_dnn}
    \end{subfigure}
    \hfill
    % ================= 4. TNDA: PAS-Net =================
    \begin{subfigure}[b]{0.24\textwidth}
        \centering
        \includegraphics[width=\linewidth]{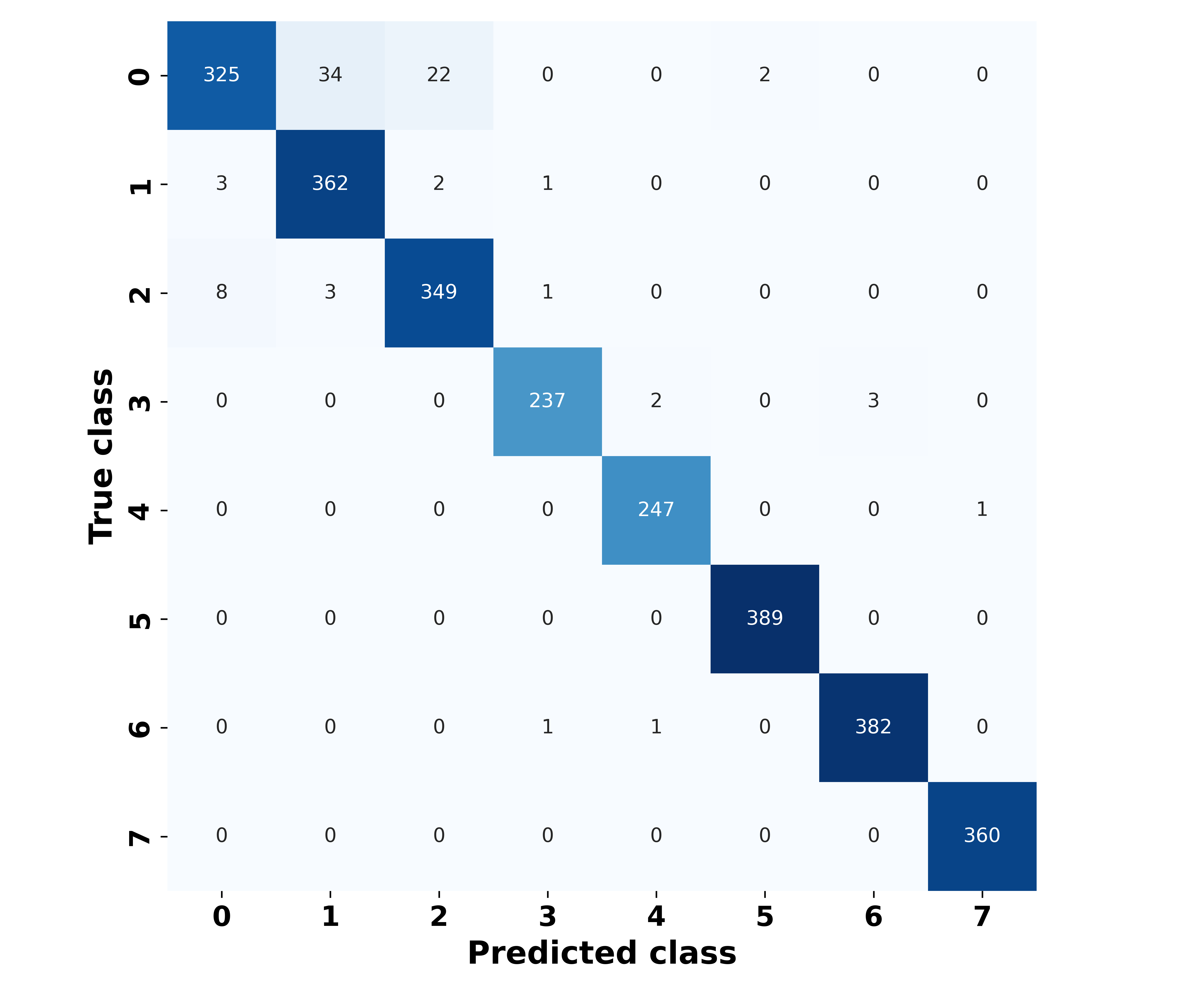}
        \caption{TNDA: PAS-Net (Ours)}
        \label{fig:cm_tnda_pasnet}
    \end{subfigure}

    \caption{Confusion matrices comparing the predictions of the baseline continuous DNN (DeepConvLSTM) against our proposed event-driven PAS-Net on the PAMAP2 and TNDA datasets. By explicitly routing physical dependencies, PAS-Net effectively suppresses off-diagonal noise and mitigates severe inter-class confusion among geometrically similar activities.}
    \label{fig:confusion_matrices}
\end{figure*}
\subsection{System Energy Efficiency and Computational Cost}

To address RQ2, we map algorithmic operations to 45nm CMOS hardware logic \cite{horowitz2014computing, panda2020toward}. Unlike power-hungry continuous DNNs relying on floating-point MACs ($4.6$ pJ), PAS-Net operates via sparse integer accumulations (AC, $0.1$ pJ). As shown in Table \ref{tab:energy_efficiency}, PAS-Net fundamentally shatters the accuracy-energy trade-off. On PAMAP2, it consumes merely $2.57~\mu$J per inference—achieving over $34\times$ energy savings compared to top DNNs (e.g., ResNet-SE at $88.24~\mu$J and DeepConvLSTM at $94.62~\mu$J) while improving absolute accuracy by $7.24\%$. This advantage scales remarkably on multi-node cohorts like TNDA-HAR, where PAS-Net ($3.23~\mu$J) requires only $1/81$ the energy of heavy attention models like Self-HAR ($263.00~\mu$J).

Within the neuromorphic domain, this efficiency stems from the extreme sparsity of internal spike trains, dynamically regulated by our Causal Neuromodulator. While unconstrained attention-based SNNs (e.g., STAtten and SpikeDrivenTransformer) suffer from a severe "spike explosion" bottleneck—generating around $200$M SOPs and consuming over $23~\mu$J on PAMAP2—PAS-Net intelligently suppresses redundant emissions to a mere $4.53$M SOPs. Crucially, PAS-Net operates with a highly compact architectural footprint of just $0.26$M parameters, yet its active power is driven even lower by this extreme event-driven sparsity rather than static weights. Thus, it perfectly strikes the Pareto front between expressive physical modeling and green computation.

Finally, we note that baselines exhibiting anomalously low FLOPs, particularly rTsfNet ($1.62~\mu$J), represent a deceptive theoretical metric. Standard profilers omit the massive CPU overhead required for rTsfNet's 45 prerequisite handcrafted features (e.g., Fast Fourier Transforms) \cite{lv2024efficient}. In stark contrast, PAS-Net is a strictly end-to-end framework directly ingesting raw IMU streams. Its measured energy reflects the true system-level cost, reinforcing its practicality for severely battery-constrained edge deployments.
% --- Master Energy Efficiency Table ---
\begin{table*}[htbp]
\centering
\caption{Comprehensive computational cost and system-level energy efficiency analysis. Network Parameters (M), theoretical FLOPs (M), and actual Synaptic Operations (SOPs, M) are profiled based on the PAMAP2 benchmark as the structural representative. The dynamic energy consumption ($\mu$J) is systematically evaluated across all seven diverse datasets to demonstrate consistent efficiency. Missing computational tracking logs from original baselines have been conservatively imputed based on average network firing rates (inclusive of the standardized FP32 continuous encoding overhead prior to the spike domain). \textbf{$\dagger$ Note:} The energy profiling for rTsfNet represents only its shallow MLP backend and strictly excludes the massive CPU-based computational overhead required to extract its 45 prerequisite handcrafted time-series features (e.g., FFT), making its true end-to-end system energy significantly higher.}
\label{tab:energy_efficiency}
\resizebox{\textwidth}{!}{
\begin{tabular}{l|ccc|ccccccc}
\toprule
\multirow{2}{*}{\textbf{Model Architecture}} & \multicolumn{3}{c|}{\textbf{Complexity (PAMAP2)}} & \multicolumn{7}{c}{\textbf{Energy Consumption ($\mu$J) across Datasets}} \\
\cmidrule(lr){2-4} \cmidrule(lr){5-11}
& \textbf{Params (M)} & \textbf{FLOPs (M)} & \textbf{SOPs (M)} & \textbf{PAMAP2} & \textbf{TNDA} & \textbf{Daily-Sports} & \textbf{HuGaDB} & \textbf{HAR70+} & \textbf{USC-HAD} & \textbf{Parkinson} \\
\midrule
DeepConvLSTM \cite{ordonez2016deep} & 0.30 & 19.31 & -- & 94.62 & 94.32 & 94.32 & 192.80 & 93.41 & 93.11 & 186.20 \\
ResNet-SE\cite{hu2018squeeze,he2016deep}& 0.82 & 18.01 & -- & 88.24 & 87.81 & 87.82 & 181.40 & 86.54 & 86.13 & 172.10 \\
MCHCNN-GRU\cite{lu2022multichannel} & 0.21 & 13.20 & -- & 64.70 & 64.24 & 64.25 & 134.80 & 62.88 & 62.44 & 124.90 \\
Self-HAR\cite{tang2021selfhar} & 0.84 & 53.79 & -- & 263.60 & 263.00 & 263.00 & 534.40 & 261.20 & 260.60 & 521.10 \\
Uni-HAR\cite{xu2023practically}& 0.84 & 53.77 & -- & 263.40 & 262.80 & 262.80 & 534.10 & 261.00 & 260.40 & 520.80 \\
rTsfNet$^\dagger$\cite{enokibori2024rtsfnet}& 0.11 & 0.35 & -- & 1.62 & 1.42 & 1.44 & 5.48 & 0.86 & 0.68 & 0.86 \\
IFConvTransformer\cite{zhang2022if}& 0.86 & 29.73 & -- & 145.70 & 145.70 & 145.70 & 291.40 & 145.60 & 145.60 & 291.20 \\
\midrule
SNN MLP & 0.01 & 3.00 & 0.18 & 1.08 & 0.91 & 0.58 & 1.38 & 0.18 & 0.18 & 0.12 \\
LMU-Former\cite{liu2024lmuformer} & 0.54 & 325.80 & 40.45 & 6.17 & 17.53 & 4.85 & 14.18 & 3.21 & 3.16 & 1.20 \\
QKFormerIMU\cite{zhou2024qkformer}& 0.09 & 54.76 & 16.19 & 2.68 & 2.73 & 1.26 & 3.46 & 0.47 & 0.62 & 0.30 \\
STAtten\cite{lee2025spiking}& 2.12 & 1278.00 & 201.60 & 24.40 & 58.62 & 16.81 & 40.50 & 12.84 & 7.76 & 1.15 \\
SpikeDrivenTransformer\cite{yao2023spike}& 2.12 & 1278.00 & 190.00 & 23.24 & 35.59 & 31.02 & 42.68 & 13.16 & 7.26 & 1.17 \\
SpikeDrivenTransformerV2\cite{yao2024spike}& 0.54 & 329.20 & 90.21 & 11.14 & 17.01 & 6.77 & 10.01 & 1.62 & 3.40 & 0.85 \\
SpikeGRU\cite{lv2024efficient}& 0.06 & 11.66 & 3.12 & 1.90 & 1.64 & 1.02 & 2.05 & 0.43 & 0.46 & 0.38 \\
SpikeRNN\cite{lv2024efficient}& 0.01 & 2.19 & 0.40 & 10.11 & 9.83 & 6.15 & 7.50 & 4.52 & 8.80 & 5.60 \\
SpikeTCN\cite{lv2024efficient}& 0.02 & 11.98 & 2.67 & 0.27 & 0.40 & 0.25 & 0.37 & 0.07 & 0.05 & 0.08 \\
Spikformer\cite{zhou2022spikformer}& 0.27 & 162.40 & 46.42 & 6.76 & 7.64 & 4.75 & 5.20 & 1.65 & 1.33 & 2.91 \\
TSSNN\cite{lv2024efficient}& 0.003 & 0.35 & 0.09 & 1.60 & 1.33 & 0.83 & 1.20 & 0.27 & 0.27 & 0.17 \\
\rowcolor{gray!10}
\textbf{PAS-Net (Ours)} & 0.26 & 40.28 & 4.53 & \textbf{2.57} & \textbf{3.23} & \textbf{2.05} & \textbf{3.89} & \textbf{0.99} & \textbf{0.87} & \textbf{1.42} \\
\bottomrule
\end{tabular}
}
\end{table*}
\begin{figure*}[t]
    \centering
    % 第一张图：静态 (Sitting)
    \begin{subfigure}[b]{0.32\textwidth}
        \centering
        \includegraphics[width=\textwidth]{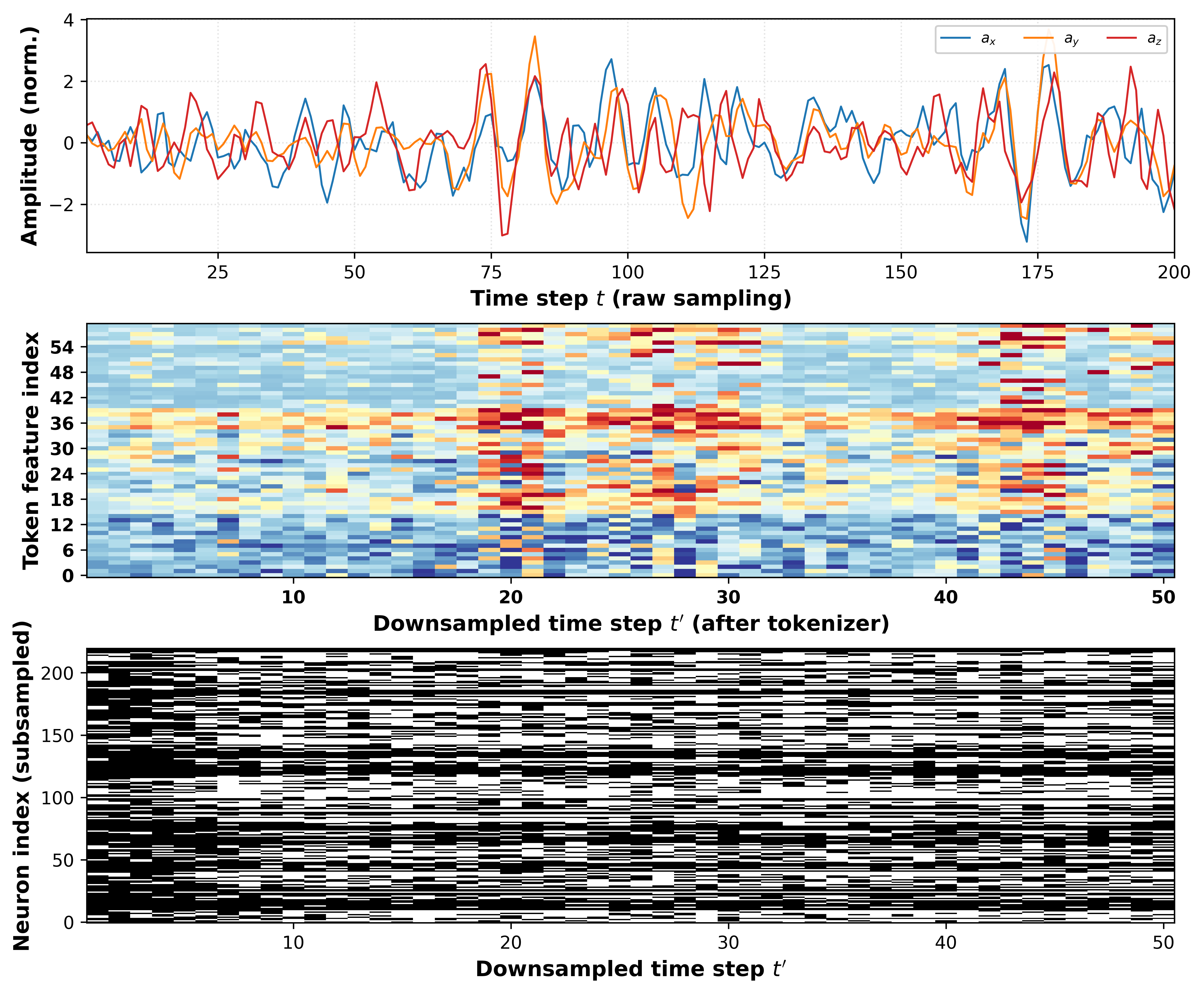}
        \caption{Sitting (TNDA)}
        \label{fig:raster_sitting}
    \end{subfigure}
    \hfill
    % 第二张图：高频动态 (Running)
    \begin{subfigure}[b]{0.32\textwidth}
        \centering
        \includegraphics[width=\textwidth]{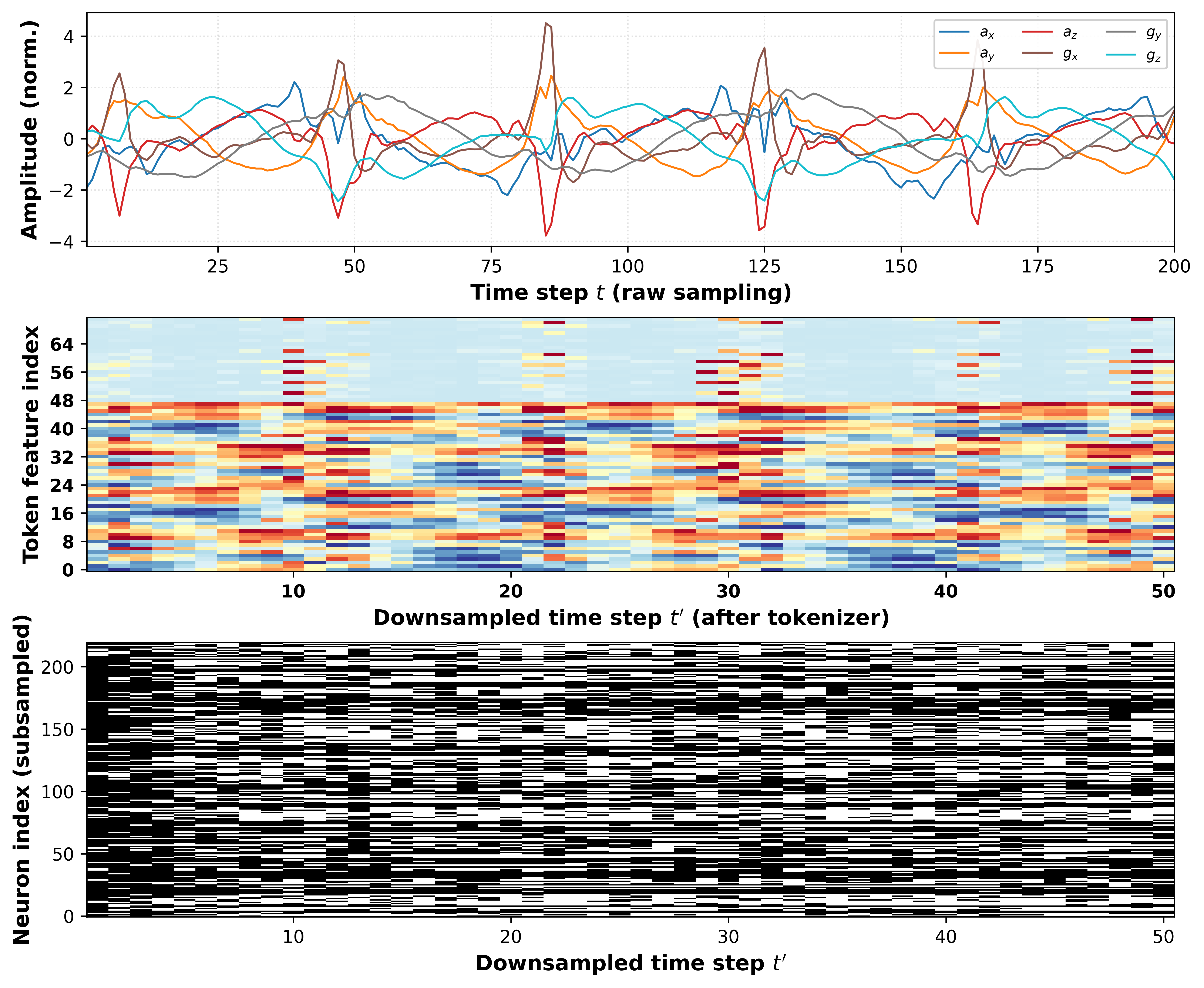}
        \caption{Running (PAMAP2)}
        \label{fig:raster_running}
    \end{subfigure}
    \hfill
    % 第三张图：复杂动态 (Ascending Stairs)
    \begin{subfigure}[b]{0.32\textwidth}
        \centering
        \includegraphics[width=\textwidth]{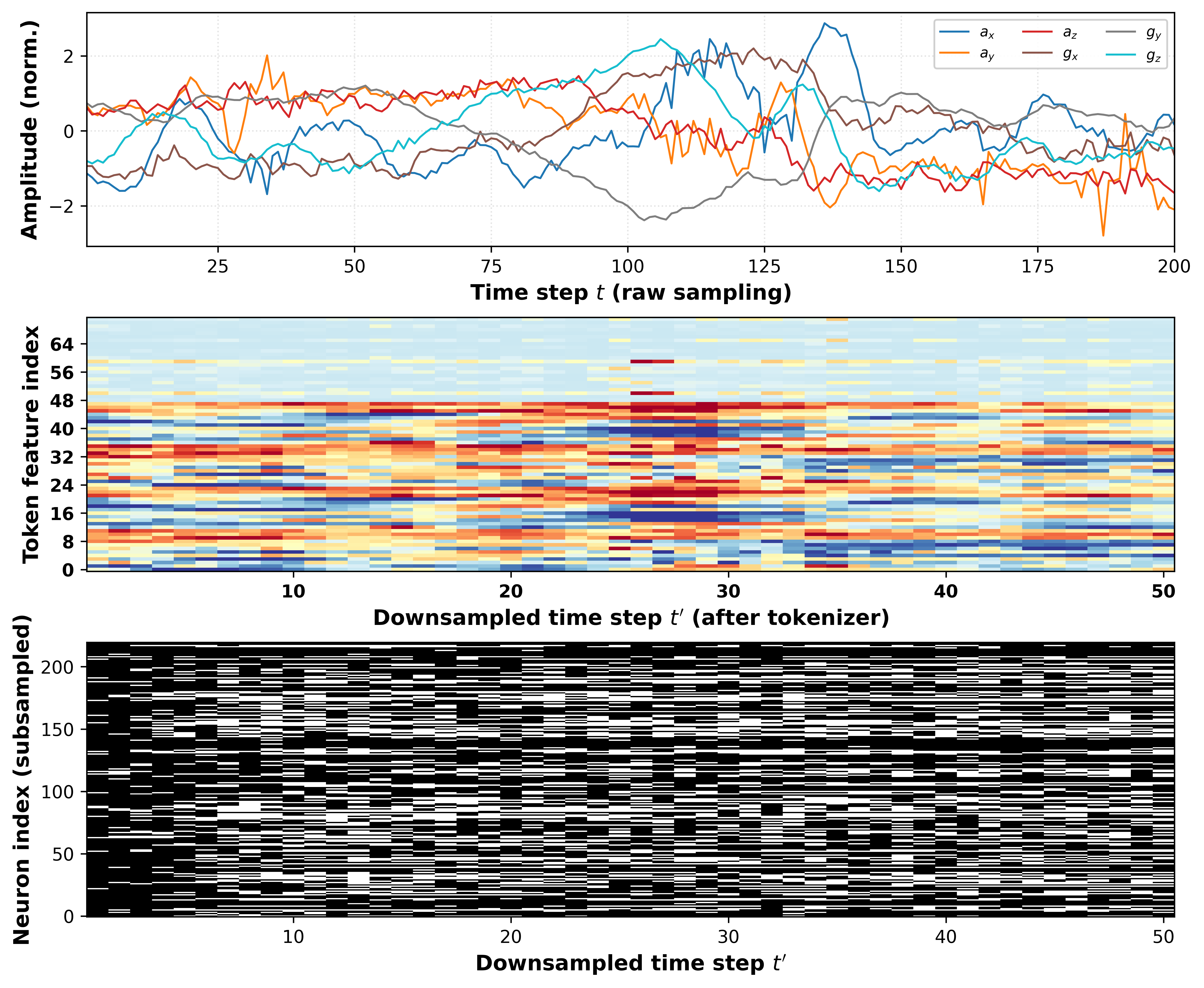}
        \caption{Ascending Stairs (PAMAP2)}
        \label{fig:raster_stairs}
    \end{subfigure}
    
    % 一句话 Caption
    \caption{Visualization of raw IMU signals (top), dense token features (middle), and deep sparse spike rasters (bottom), demonstrating the extreme sparsity and distinct spatiotemporal firing patterns learned by PAS-Net.}
    \label{fig:spike_raster_comparison}
\end{figure*}
As illustrated in Figure \ref{fig:spike_raster_comparison}, we visualize the signal processing hierarchy of PAS-Net across different activity classes. A vertical comparison reveals that the network successfully compresses noisy, dense continuous signals into highly sparse, event-driven spike trains, leaving large silent regions that physically validate its extreme energy efficiency. Furthermore, a horizontal comparison demonstrates distinct spatiotemporal fingerprints: static postures like \textit{Sitting} (Fig. \ref{fig:raster_sitting}) exhibit minimal and uniform activations, while dynamic activities such as \textit{Running} (Fig. \ref{fig:raster_running}) and \textit{Ascending Stairs} (Fig. \ref{fig:raster_stairs}) induce rhythmic, burst-firing patterns strongly aligned with critical motion phases. This confirms that PAS-Net learns task-dependent neural barcodes rather than random sparsity, effectively filtering out irrelevant background noise while preserving crucial motion kinematics.
\subsection{System Latency and Early-Exit Dynamics}
\label{sec:early_exit}

\begin{figure*}[htbp] % 双栏浮动体
    \centering
    % ================= 左侧：折线图 =================
    \begin{minipage}[b]{0.52\textwidth} % [b] 表示底部对齐
        \centering
        \includegraphics[width=\linewidth]{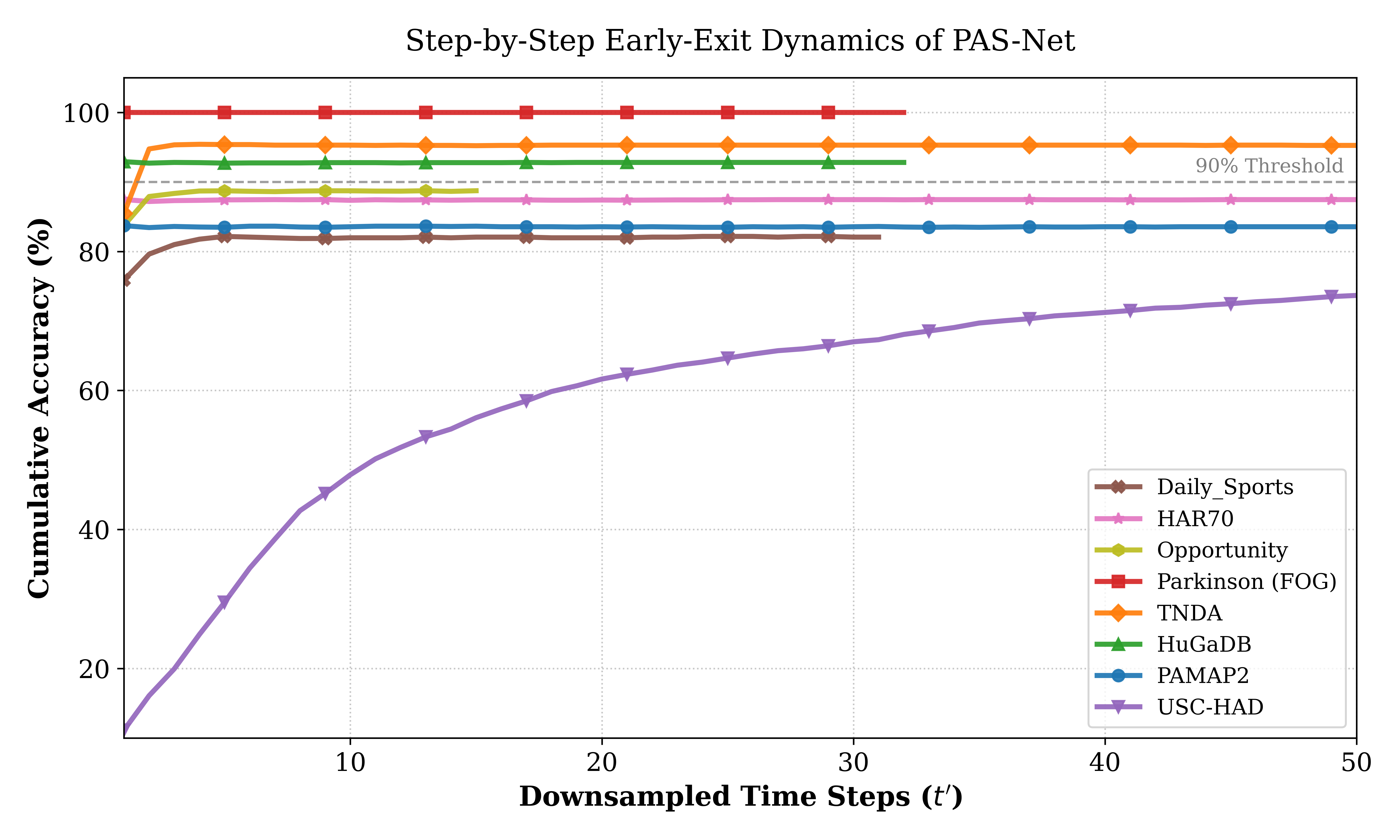}
        
        \vspace{0.1cm} % 微调间距，使其与右侧对齐感更好
        \caption{Step-by-step cumulative accuracy ($t'$) tracking the early-exit inference dynamics of PAS-Net. Structurally explicit activities trigger sub-second early exits, while complex transitions accumulate evidence over the full horizon.}
        \label{fig:early_exit_dynamics}
    \end{minipage}
    \hfill % 撑开中间的空白
    % ================= 右侧：量化数据表 =================
    \begin{minipage}[b]{0.45\textwidth} % 适当调整宽度，避免表格被压扁
        \centering
        \resizebox{\linewidth}{!}{
        \begin{tabular}{lccc}
        \toprule
        \textbf{Dataset} & \textbf{Total $T'$} & \textbf{Exit Step ($t'_{exit}$)} & \textbf{\shortstack{Dynamic Energy \\ Saved (\%)}} \\
        \midrule
        \textbf{HAR70}     & 50 & 1 & \textbf{98.0\%} \\
        \textbf{PAMAP2}    & 50 & 1 & \textbf{98.0\%} \\
        \textbf{Parkinson} & 32 & 1 & \textbf{96.9\%} \\
        \textbf{HuGaDB}    & 32 & 1 & \textbf{96.9\%} \\
        \textbf{TNDA}      & 50 & 3 & \textbf{94.0\%} \\
        \textbf{Daily-Sports}& 31 & 4 & \textbf{87.1\%} \\
        \textbf{Opportunity}& 15 & 4 & \textbf{73.3\%} \\
        \textbf{USC-HAD}   & 50 & 49 & 2.0\% \\
        \bottomrule
        \end{tabular}
        }
        
        \vspace{0.3cm} % 给表格和 Caption 之间留出呼吸感
        % 注意：这里把 captionof 放在了 tabular 之后！
        \captionof{table}{Quantitative dynamic energy reduction via early-exit mechanism. Metrics are reported at the earliest time step ($t'_{exit}$) where the confidence strictly satisfies a \textbf{99.5\% relative peak accuracy} threshold.}
        \label{tab:early_exit_savings}
    \end{minipage}
\end{figure*}

To address RQ3, we investigate PAS-Net's real-time responsiveness. By continuously evaluating dense logits via the Temporal Spike Error (TSE) loss, PAS-Net shatters the traditional ``buffer-and-compute'' latency deadlock, dynamically halting computation once a 99.5\% relative peak accuracy threshold is reached. As illustrated in Fig. \ref{fig:early_exit_dynamics} and Table \ref{tab:early_exit_savings}, this input-adaptive paradigm yields profound task-dependent dynamics. Highly distinct multi-node or pathological patterns (e.g., Parkinson, PAMAP2, HAR70) trigger instantaneous convergence at $t'=1$, leveraging the causal neuromodulator to rapidly accumulate global evidence and saving up to 98.0\% of dynamic energy. Conversely, ambiguous single-node transitions (e.g., USC-HAD) naturally prompt cautious evidence accumulation over the full temporal horizon ($t'=49$). Crucially, this step-by-step sequential evaluation incurs zero causality or latency penalties from the Temporal-Aware Batch Normalization (T-BN) layer, as its frozen running statistics are mathematically folded into the preceding convolutional weights prior to edge inference. Ultimately, PAS-Net fluidly scales its cognitive effort based on physical complexity, offering a pristine, Zero-MAC trade-off between ultra-low latency and peak precision.
\subsection{Ablation Study}
\label{sec:ablation}

\begin{figure*}[htbp]
    \centering
    % ================= 左侧：IMUZero风格的分组柱状图 =================
    \begin{minipage}[b]{0.48\textwidth}
        \centering
        % 确保图片放在 picture 文件夹下，且文件名对应
        \includegraphics[width=\linewidth]{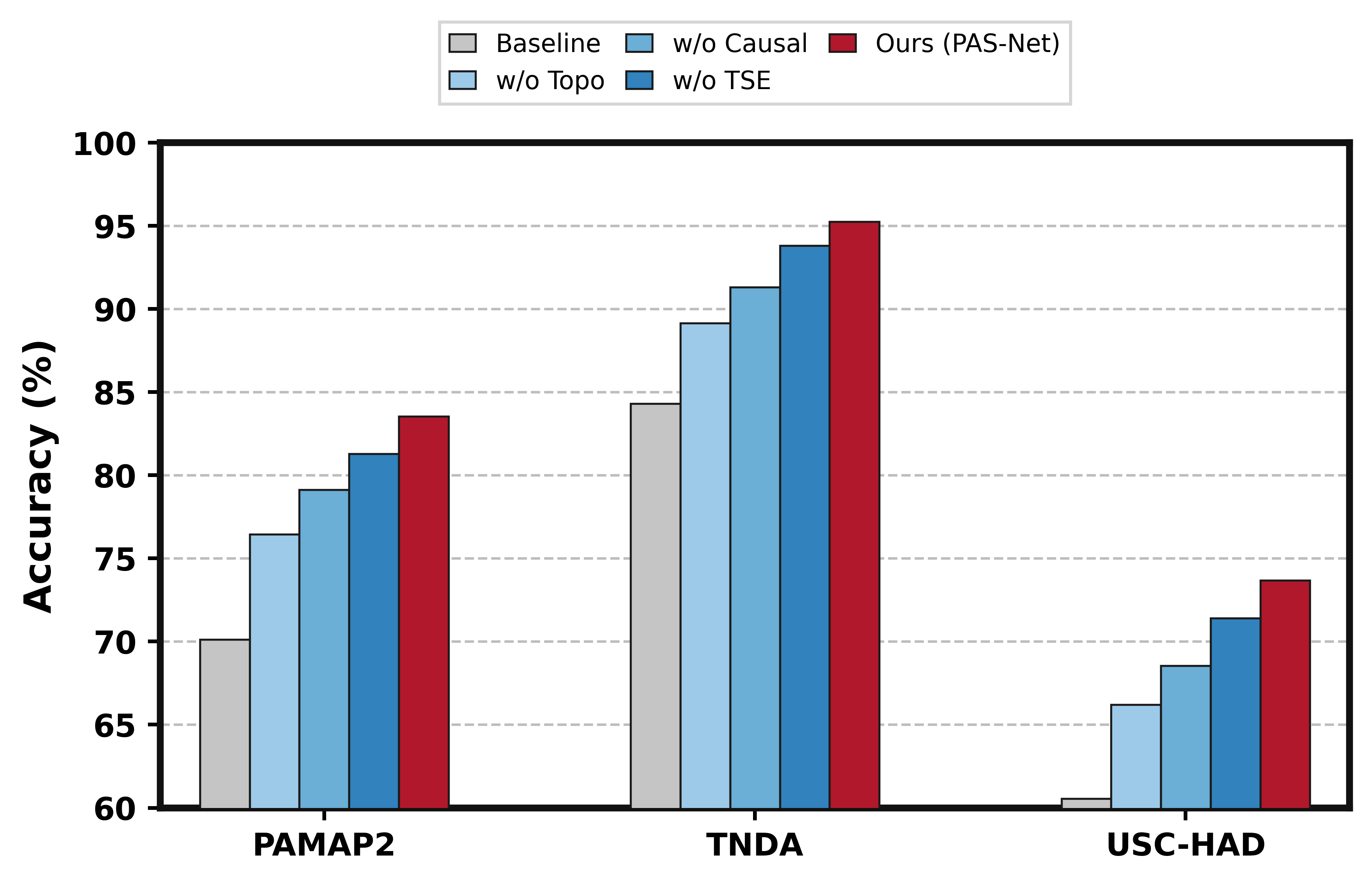} 
        \caption{Ablation study on key components of PAS-Net across three datasets. 'w/o' denotes the removal or standard replacement of the specific module.}
        \label{fig:ablation_chart}
    \end{minipage}
    \hfill
    % ================= 右侧：严谨的打勾量化表格 =================
    \begin{minipage}[b]{0.48\textwidth}
        \centering
        \resizebox{\linewidth}{!}{
        \begin{tabular}{ccc|ccc}
        \toprule
        \textbf{Topo} & \textbf{Causal} & \textbf{TSE} & \textbf{PAMAP2} & \textbf{TNDA} & \textbf{USC-HAD} \\
        \midrule
        % 纯基线版本 (Baseline: 普通SNN)
        $\times$ & $\times$ & $\times$ & 70.12 & 84.30 & 60.55 \\ 
        \midrule % 顶会细节：用一条细线把基线和模型变体隔开
        $\times$ & $\checkmark$ & $\checkmark$ & 76.45 & 89.15 & 66.20 \\ % w/o Topo
        $\checkmark$ & $\times$ & $\checkmark$ & 79.12 & 91.30 & 68.55 \\ % w/o Causal
        $\checkmark$ & $\checkmark$ & $\times$ & 81.30 & 93.80 & 71.40 \\ % w/o TSE
        \rowcolor{gray!10}
        $\checkmark$ & $\checkmark$ & $\checkmark$ & \textbf{83.55} & \textbf{95.25} & \textbf{73.67} \\ % PAS-Net 完整版
        \bottomrule
        \end{tabular}
        }
        \vspace{0.3cm} 
        \captionof{table}{Quantitative ablation results (Accuracy \%). The baseline model ($\times, \times, \times$) represents a standard multi-layer SNN without our proposed adaptive and temporal mechanisms.}
        \label{tab:ablation_table}
    \end{minipage}
\end{figure*}

To provide deeper insight into the architectural superiority of PAS-Net, we conduct a comprehensive ablation study to evaluate the core contributions of each proposed component: 1) the Adaptive Symmetric Topology Routing (Topo); 2) the $O(1)$ Causal Neuromodulation (Causal); and 3) the Temporal Spike Error loss (TSE). We systematically evaluate these variants across three representative datasets. The quantitative results and visual comparisons are presented in Table~\ref{tab:ablation_table} and Figure~\ref{fig:ablation_chart}, respectively.

As illustrated in Figure~\ref{fig:ablation_chart} and quantitatively detailed in Table~\ref{tab:ablation_table}, each proposed component contributes significantly to the final performance of PAS-Net.

\textbf{Effectiveness of Adaptive Symmetric Topology Routing.} The removal of the spatial topology mixer (w/o Topo) leads to the most substantial performance degradation, particularly on multi-node datasets such as TNDA (dropping from 95.25\% to 89.15\%) and PAMAP2. This steep decline validates our hypothesis that treating physical sensor nodes as isolated channels results in severe spatial blindness. By explicitly enforcing a symmetric adjacency mask, the network successfully routes and aggregates physical joint dependencies, which is crucial for distinguishing geometrically similar activities.

\textbf{Impact of Causal Neuromodulation.} Replacing the causal EMA neuromodulator with a standard fixed-threshold LIF neuron (w/o Causal) significantly impairs accuracy across all benchmarks. Human activities are highly non-stationary, alternating between static postures and explosive bursts. A static threshold struggles to adapt to these shifting energy envelopes, leading to either missing features or excessive noisy spikes. The $O(1)$ causal modulation dynamically scales the firing sensitivity, acting as an event-driven filter that preserves critical kinematics while suppressing irrelevant background noise.

\textbf{Contribution of Temporal Spike Error (TSE).} While the primary objective of the TSE loss is to unlock sub-second early exits, ablating it (w/o TSE) and relying solely on a final-step Cross-Entropy loss also degrades the overall peak accuracy. This indicates that the dense temporal supervision provided by TSE effectively mitigates the gradient degradation problem commonly observed when training deep SNNs over long sequential windows via BPTT, ensuring more robust and stable spatotiemporal feature learning.
\subsection{Latent Space Visualization}

\label{sec:tsne}

\begin{figure*}[htbp]
    \centering
    % 第一行：经典基线 vs. 复杂模型失效
    \begin{subfigure}[b]{0.23\textwidth}
        \centering
        \includegraphics[width=\linewidth]{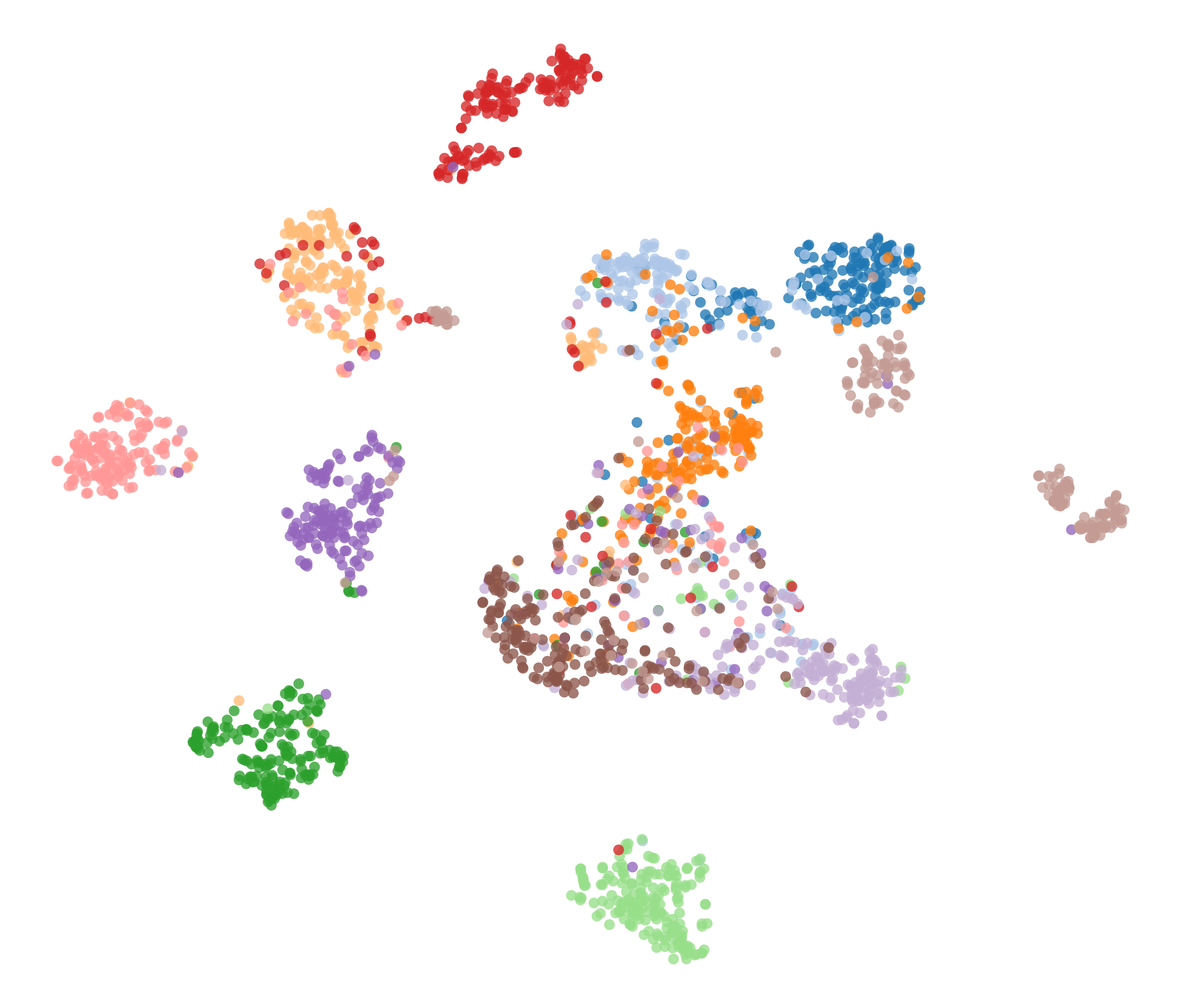}
        \caption{Continuous DNN (DeepConvLSTM)}
        \label{fig:tsne_dnn}
    \end{subfigure}
    \hfill
    \begin{subfigure}[b]{0.23\textwidth}
        \centering
        \includegraphics[width=\linewidth]{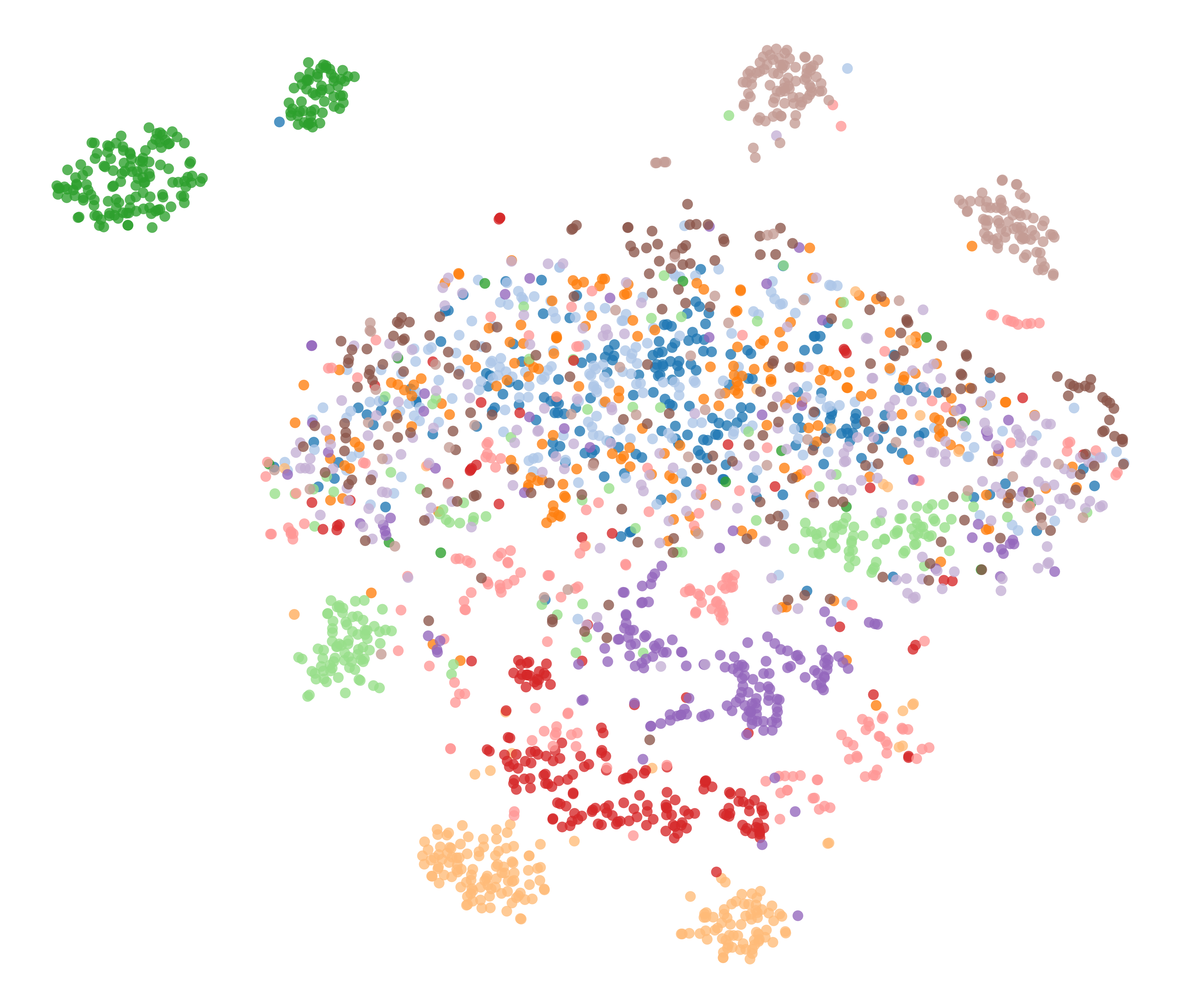}
        \caption{Advanced Baseline (LMUFormer)}
        \label{fig:tsne_lmuformer}
    \end{subfigure}
 \begin{subfigure}[b]{0.23\textwidth}
        \centering
        \includegraphics[width=\linewidth]{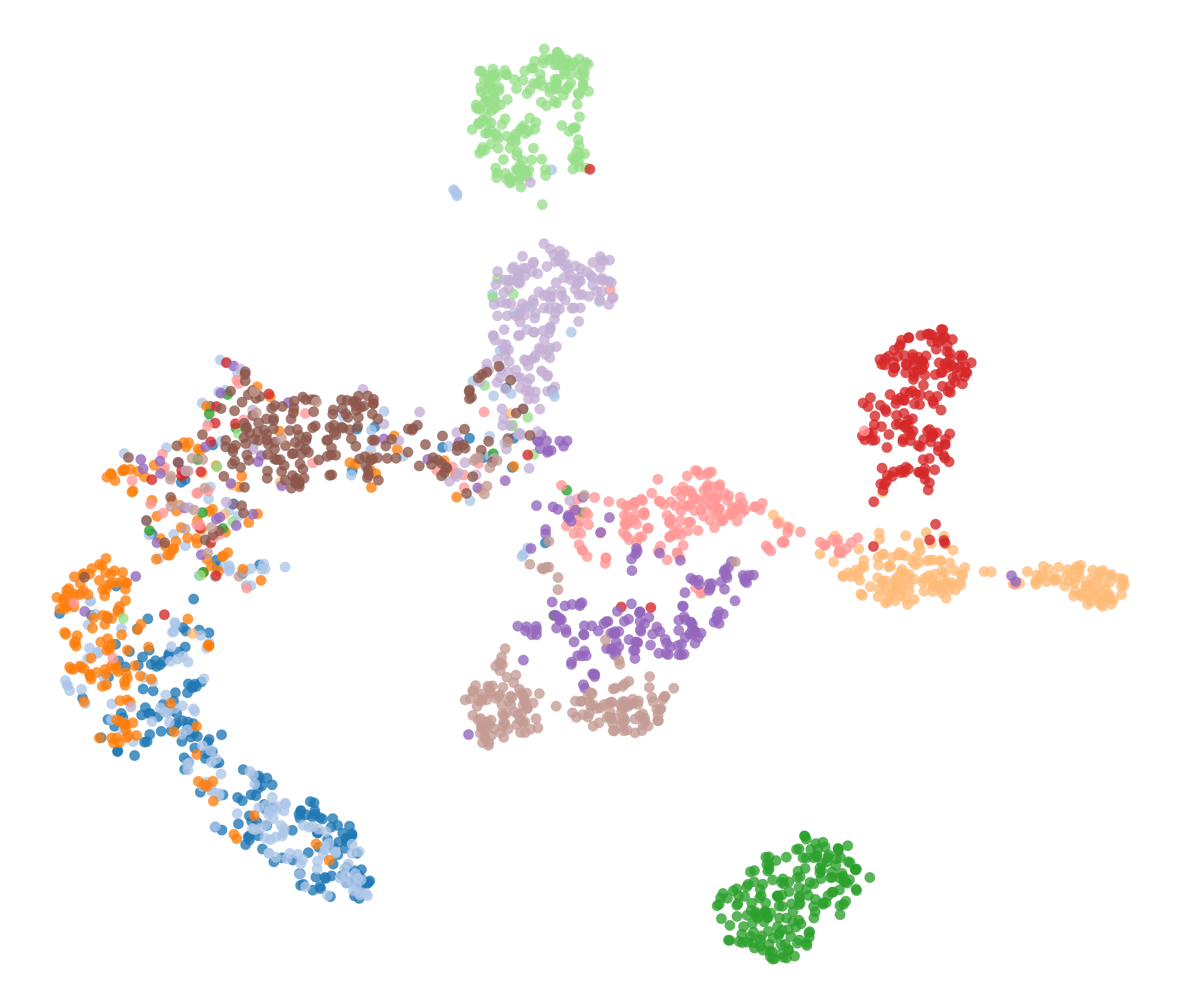}
        \caption{PAS-Net (w/o Causal Neuromodulation)}
        \label{fig:tsne_lite}
    \end{subfigure}
    \hfill
    \begin{subfigure}[b]{0.23\textwidth}
        \centering
        \includegraphics[width=\linewidth]{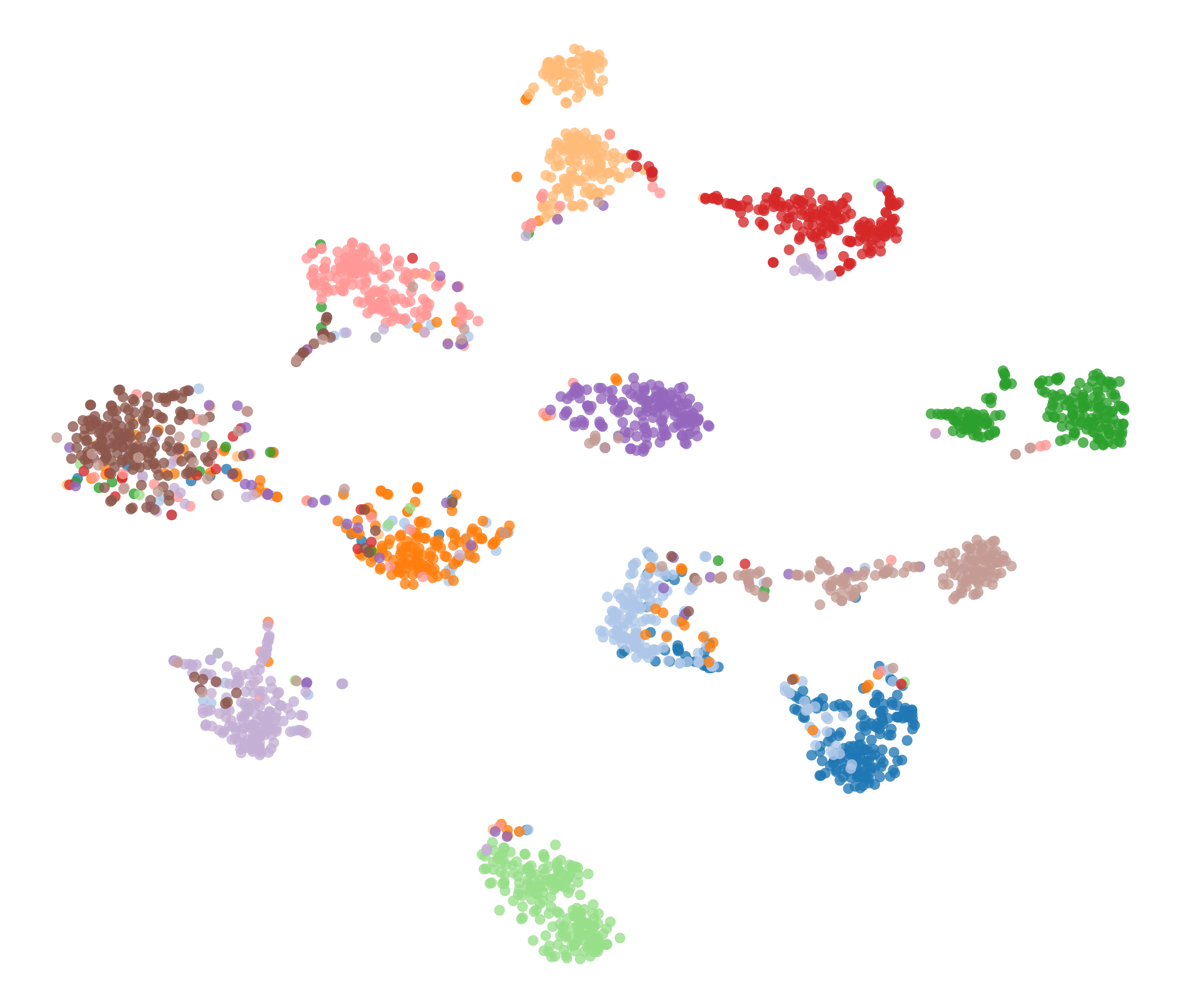}
        \caption{Proposed PAS-Net (Full)}
       \label{fig:tsne_full}
    \end{subfigure}
    \caption{t-SNE visualization of the learned latent spaces on the PAMAP2 dataset. The progression from (a) and (b) to (d) illustrates how our explicit spatial topology routing and temporal causal neuromodulation systematically disentangle overlapping kinematic signatures, culminating in a highly structured and separable feature space.}

    \label{fig:tsne_evolution}

\end{figure*}

To qualitatively demonstrate the evolutionary superiority of the representations learned by PAS-Net, we employ t-SNE to project the high-dimensional features from the penultimate layer into a 2D latent space. Figure~\ref{fig:tsne_evolution} visualizes the feature distributions across four different architectural paradigms on the PAMAP2 dataset.

As observed in Figure~\ref{fig:tsne_evolution}(a), the standard continuous DNN (DeepConvLSTM) establishes baseline clusters but struggles with severe inter-class overlap in the central manifold. More severely, as shown in Figure~\ref{fig:tsne_evolution}(b), directly applying advanced temporal architectures like LMUFormer without physical constraints results in a collapsed, indistinguishable latent space, highlighting the unsuitability of naive sequence models for raw IMU streams.

In contrast, incorporating our proposed spatial mechanisms (Figure~\ref{fig:tsne_evolution}c) begins to clearly isolate physical node dependencies, though some decision boundaries remain fuzzy. Ultimately, the complete PAS-Net equipped with causal neuromodulation (Figure~\ref{fig:tsne_evolution}d) generates a highly structured, separable, and compact latent space. It effectively filters out shared background noise, proving the necessity of dual-adaptive spiking dynamics for precise activity disentanglement.
\section{Discussion and Limitations}
\label{sec:discussion}

While PAS-Net demonstrates state-of-the-art performance and unprecedented energy efficiency for continuous IMU-based HAR, several avenues remain for future exploration to bridge the gap toward ubiquitous real-world deployment.

\textbf{Deployment on Physical Neuromorphic Hardware.} In this work, the energy consumption of PAS-Net is quantitatively estimated based on well-established 45nm CMOS hardware logic, strictly tracking the actual Synaptic Operations (SOPs). While this is a standard and rigorous evaluation metric in algorithmic neuromorphic research, the ultimate goal is to deploy PAS-Net on specialized neuromorphic chips (e.g., Intel Loihi or IBM TrueNorth). Physical deployment involves translating the algorithmic causal neuromodulation and temporal-wise exits into hardware-specific routing protocols. Exploring the on-chip latency and power consumption of PAS-Net on such asynchronous hardware remains our immediate future work.

\textbf{Robustness to Sensor Displacement.} PAS-Net utilizes an Adaptive Symmetric Topology mixer to explicitly route physical joint dependencies. However, in unconstrained daily scenarios, users rarely wear devices at the exact same location or orientation as in the training protocols (e.g., a smartwatch sliding down the wrist). Although our invariant tokenizer mitigates orientation shifts to some extent, severe topological displacement may perturb the pre-learned spatial adjacency matrix. Future iterations could integrate unsupervised domain adaptation or dynamic graph learning to autonomously calibrate the topology mask against on-the-fly sensor shifts.

\textbf{On-Device Continual Learning.} Currently, PAS-Net is trained offline and deployed for inference. Wearable devices generate massive amounts of personalized, unlabeled data daily. The event-driven sparsity and multiplier-free nature of PAS-Net make it an ideal candidate for ultra-low-power on-device learning. Investigating Spike-Timing-Dependent Plasticity (STDP) or other biologically plausible local learning rules to enable PAS-Net to continually adapt to a specific user's evolving motion habits without cloud intervention is a highly promising direction.

\section{Conclusion}
\label{sec:conclusion}

In this paper, we identify and tackle the critical arithmetic-latency deadlock that plagues modern Deep Neural Networks in wearable sensing. We introduce the Physics-Aware Spiking Neural Network (PAS-Net), pioneering a fully multiplier-free, event-driven paradigm tailored for Green Human Activity Recognition. By synergizing an Adaptive Symmetric Topology mixer with $O(1)$-memory Causal Neuromodulation, PAS-Net successfully endows spiking networks with the capability to capture explicit biomechanical dependencies and adapt to non-stationary motion rhythms. Furthermore, leveraging a continuous Temporal Spike Error objective, we unlock an unprecedented sub-second early-exit mechanism. 

Extensive evaluations across seven diverse IMU datasets demonstrate that PAS-Net not only matches or surpasses the recognition accuracy of heavy continuous DNNs but fundamentally shatters the energy-latency barrier. It slashes the operational energy footprint to the magnitude of 0.1 pJ integer additions (saving over 90\% dynamic energy) and reduces decision latency by up to 70\% via confidence-driven early exits. Ultimately, PAS-Net establishes a highly robust and viable standard for the next generation of always-on, ultra-low-power wearable computing.
\bibliographystyle{ACM-Reference-Format}
\bibliography{sample-base}

\appendix
\section{Detailed Network Architecture and Hyperparameters}
\label{app:architecture}

To ensure the rigorous reproducibility of our proposed framework, we provide the explicit internal tensor transformations and dataset-specific hyperparameter configurations.

Table~\ref{tab:tensor_flow} details the exact layer-wise configurations and the spatiotemporal tensor flow of PAS-Net, taking the highly complex PAMAP2 dataset as a structural representative. Note that the batch size dimension is omitted for brevity. A critical design choice in our architecture is the dimension permutation after the Invariant Tokenizer. The network explicitly transitions into a time-first tensor layout $[T', C, V]$, which strictly aligns with the event-driven Spiking Neural Network backend and mathematically optimizes the causal recurrent computations inside the dynamic LIF neurons.

\begin{table}[h]
\centering
\caption{Detailed architectural configuration and internal tensor flow of PAS-Net (exemplified on the PAMAP2 dataset). The tensor shapes follow the $[T, C, V]$ layout, where $T$ is the temporal length, $C$ is the feature dimension, and $V$ is the number of physical sensor nodes. For PAMAP2, the initial temporal window $T=200$ is robustly downsampled to $T'=50$, and $V=3$.}
\label{tab:tensor_flow}
\resizebox{0.95\linewidth}{!}{
\begin{tabular}{llcl}
\toprule
\textbf{Stage} & \textbf{Module Component} & \textbf{Output Tensor Shape} & \textbf{Configuration Details} \\
\midrule
\textbf{Input} & Raw Continuous IMU Stream & $[200, 6, 3]$ & $C_{in}=6$ (Acc + Gyro), $V=3$ spatial nodes \\
\midrule
\multirow{2}{*}{\textbf{Perceptual Frontend}} 
& Invariant Tokenizer & $[50, 24, 3]$ & Extract mean, max, var, $L_2$ norm (stride=4) \\
& Permute \& Spiking Embed & $[50, 256, 3]$ & Conv1d ($24 \to 256$), T-BN, Dynamic LIF \\
\midrule
\multirow{2}{*}{\textbf{Deep Spiking Core}} 
& \multirow{2}{*}{PAS-Block $\times L$} & \multirow{2}{*}{$[50, 256, 3]$} & Stage 1: Symmetric Mask $\tilde{\mathbf{A}}$, Spatiotemporal Conv ($k=5$) \\
& & & Stage 2: Spiking Dilated TCN, $D=256$, Depth $L=5$ \\
\midrule
\multirow{2}{*}{\textbf{Readout \& Exit}} 
& Spatial Pooling & $[50, 256]$ & Fused Mean + Max Pooling across the $V$ dimension \\
& Linear Classifier & $[50, 12]$ & Fully Connected Layer mapping to $K=12$ classes \\
\bottomrule
\end{tabular}
}
\end{table}

Furthermore, human activities inherently exhibit varying degrees of biomechanical complexity and temporal periodicity across different datasets. To optimally accommodate these variations, PAS-Net employs a frequency-adaptive hyperparameter scaling strategy. As summarized in Table~\ref{tab:hyperparams}, network depth ($L$), embedding dimensions ($D$), and MLP expansion ratios are dynamically scaled. For instance, the Daily-Sports dataset encompasses full-body, long-duration athletic movements (e.g., rowing, playing basketball). Consequently, we significantly expand its network depth to $L=7$ and the MLP ratio to $4.0$ to construct deeper temporal receptive fields and capture highly complex kinetic representations.

\begin{table}[h]
\centering
\caption{Dataset-specific core hyperparameter configurations for PAS-Net. }
\label{tab:hyperparams}
\resizebox{0.9\linewidth}{!}{
\begin{tabular}{lcccccc}
\toprule
\textbf{Dataset} & \textbf{Nodes ($V$)} & \textbf{Input Ch. ($C_{in}$)} & \textbf{Classes ($K$)} & \textbf{Embed Dim ($D$)} & \textbf{Depth ($L$)} & \textbf{MLP Ratio} \\
\midrule
\textbf{PAMAP2} & 3 & 6 & 12$^\dagger$ & 128 & 3 & 2.0 \\
\textbf{Daily-Sports} & 5 & 3 & 19 & 192 & 7 & 4.0 \\
\textbf{TNDA-HAR} & 5 & 3 & 8 & 128 & 5 & 3.0 \\
\textbf{HuGaDB} & 5 & 6 & 12 & 128 & 5 & 3.0 \\
\textbf{USC-HAD} & 1 & 6 & 12 & 128 & 5 & 3.0 \\
\textbf{HAR70+} & 2 & 3 & 7 & 96 & 4 & 2.5 \\
\textbf{Parkinson (FOG)} & 1 & 3 & 2 & 128 & 5 & 3.0 \\
\bottomrule
\end{tabular}
}
\vspace{0.1cm}
\small{\\$^\dagger$\textit{Note:} For PAMAP2, we strictly adhere to the standard rigorous protocol by evaluating on the 12 salient physical activities, omitting unstructured background and transient classes from the classification head.}
\end{table}
\section{Visualization of Layer-wise Physical Topology Evolution}
\label{app:learned_topology}

A core contribution of PAS-Net is the Adaptive Symmetric Topology routing, designed to inherently capture biomechanical linkages. To demonstrate that this module genuinely learns meaningful physical correlations across different scales, we conduct a comprehensive case study visualizing the learned symmetric adjacency masks ($\tilde{\mathbf{A}}$) from three representative multi-node datasets: PAMAP2 ($V=3$), Daily-Sports ($V=5$), and HuGaDB ($V=6$).

\begin{figure*}[htbp]
    \centering
    % 第一张图：PAMAP2 (3个节点)
    \begin{subfigure}[b]{\textwidth}
        \centering
        \includegraphics[width=\linewidth]{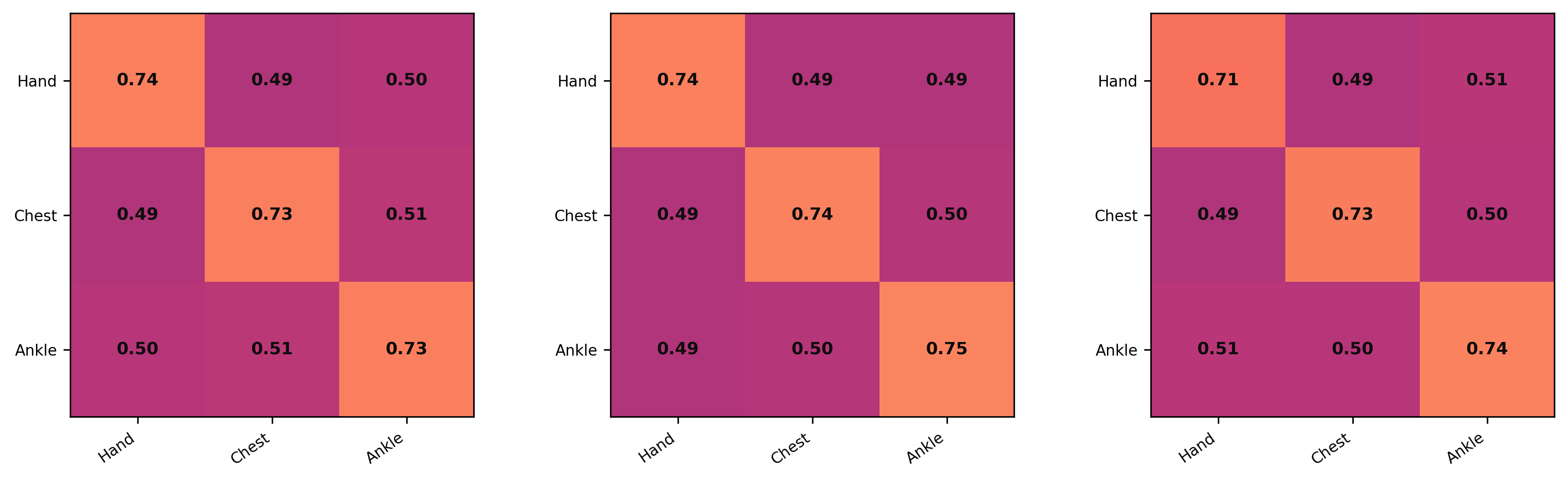}
        \vspace{-0.5cm} % 微调间距
        \caption{PAMAP2 ($V=3$): Sparse full-body topology (Hand, Chest, Ankle).}
        \label{fig:topo_pamap2}
    \end{subfigure}
    
    \vspace{0.4cm} % 组间距
    
    % 第二张图：Daily-Sports (5个节点)
    \begin{subfigure}[b]{\textwidth}
        \centering
        \includegraphics[width=\linewidth]{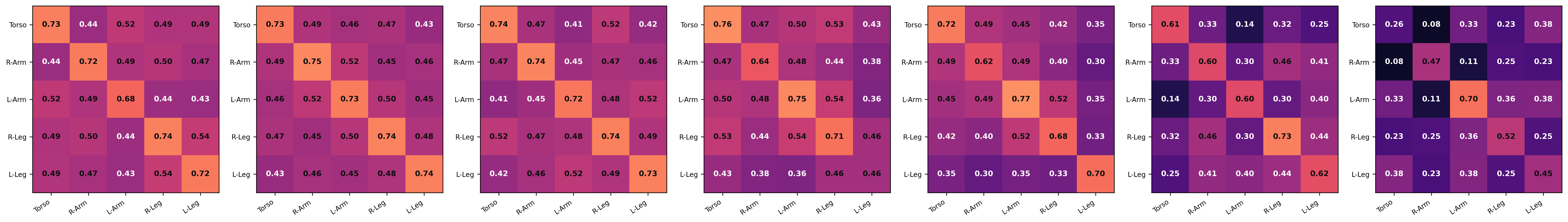}
        \vspace{-0.5cm}
        \caption{Daily-Sports ($V=5$): Standard full-body extremities and torso topology.}
        \label{fig:topo_dailysports}
    \end{subfigure}
    
    \vspace{0.4cm}
    
    % 第三张图：HuGaDB (6个节点)
    \begin{subfigure}[b]{\textwidth}
        \centering
        \includegraphics[width=\linewidth]{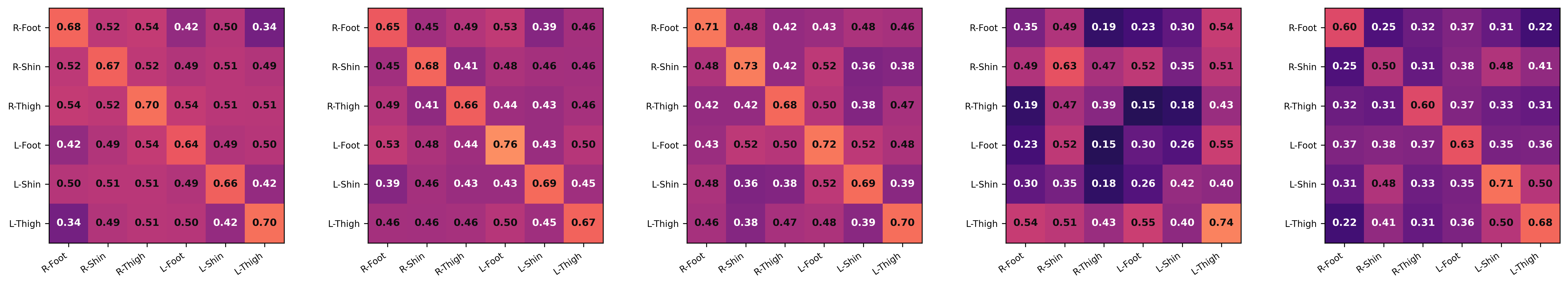}
        \vspace{-0.5cm}
        \caption{HuGaDB ($V=6$): Dense bilateral lower-body topology for highly symmetric gait analysis.}
        \label{fig:topo_hugadb}
    \end{subfigure}
    
    \caption{Layer-wise heatmap visualization of the learned symmetric spatial topology masks ($\tilde{\mathbf{A}}$) across the stacked PAS-Blocks. From left to right, the matrices represent the spatial routing weights from shallow to deep layers. The explicit mathematical constraint ensures strict bidirectional symmetry ($\tilde{\mathbf{A}} = \tilde{\mathbf{A}}^\top$), effectively mimicking human joint physical equivalence.}
    \label{fig:topology_evolution}
\end{figure*}

Crucially, rather than presenting a single static mask, Figure~\ref{fig:topology_evolution} visualizes the \textbf{layer-wise topological evolution} across the stacked PAS-Blocks (from shallow layers on the left to deep layers on the right).

As illustrated in Figure~\ref{fig:topology_evolution}, the learned spatial masks exhibit two profound physical insights:
\begin{enumerate}
    \item \textbf{Biomechanical Alignment:} The high activation off-diagonal weights perfectly align with actual human kinetic chains. For instance, in HuGaDB (Figure~\ref{fig:topo_hugadb}), the bilateral symmetry between the left and right lower limbs is distinctly captured, reflecting the periodic nature of human gait. In Daily-Sports (Figure~\ref{fig:topo_dailysports}), strong reciprocal attention is established between the four extremities and the central Torso.
    \item \textbf{Layer-wise Receptive Expansion:} From shallow to deep PAS-Blocks (left to right), the network fundamentally shifts its attention. Shallow layers exhibit strong diagonal dominance, indicating localized self-feature extraction. In contrast, deeper layers progressively activate off-diagonal elements, actively routing and fusing global spatiotemporal kinematics across distant joints to construct high-level semantic representations.
\end{enumerate}
\section{Micro-Dissection of Temporal Firing Dynamics}
\label{app:temporal_dynamics}

\begin{figure*}[htbp]
    \centering
    % 左图：Walking
    \begin{subfigure}[b]{0.48\textwidth}
        \centering
        \includegraphics[width=\linewidth]{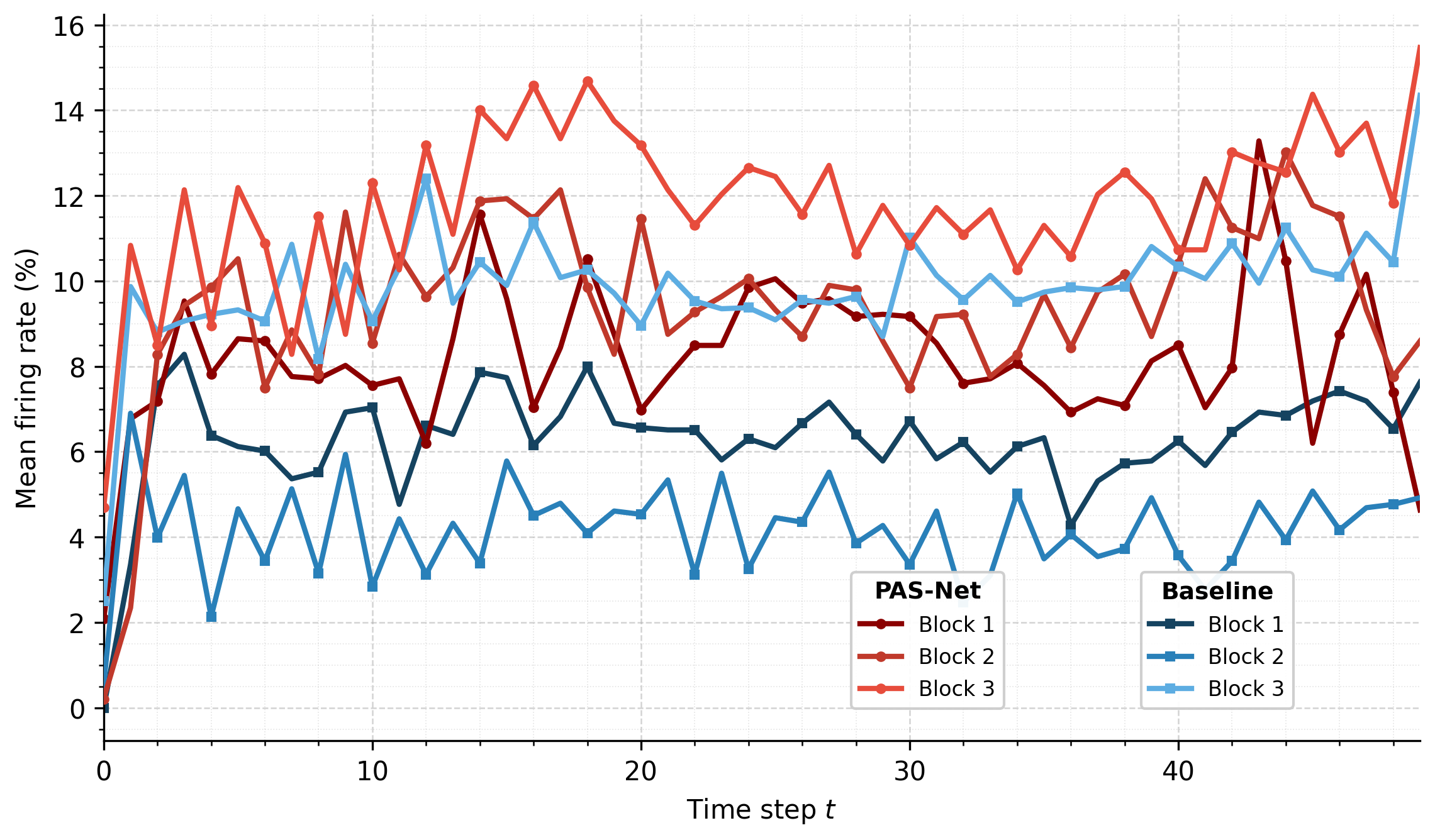}
        \caption{Walking sequence}
        \label{fig:valve_walking}
    \end{subfigure}
    \hfill
    % 右图：Running
    \begin{subfigure}[b]{0.48\textwidth}
        \centering
        \includegraphics[width=\linewidth]{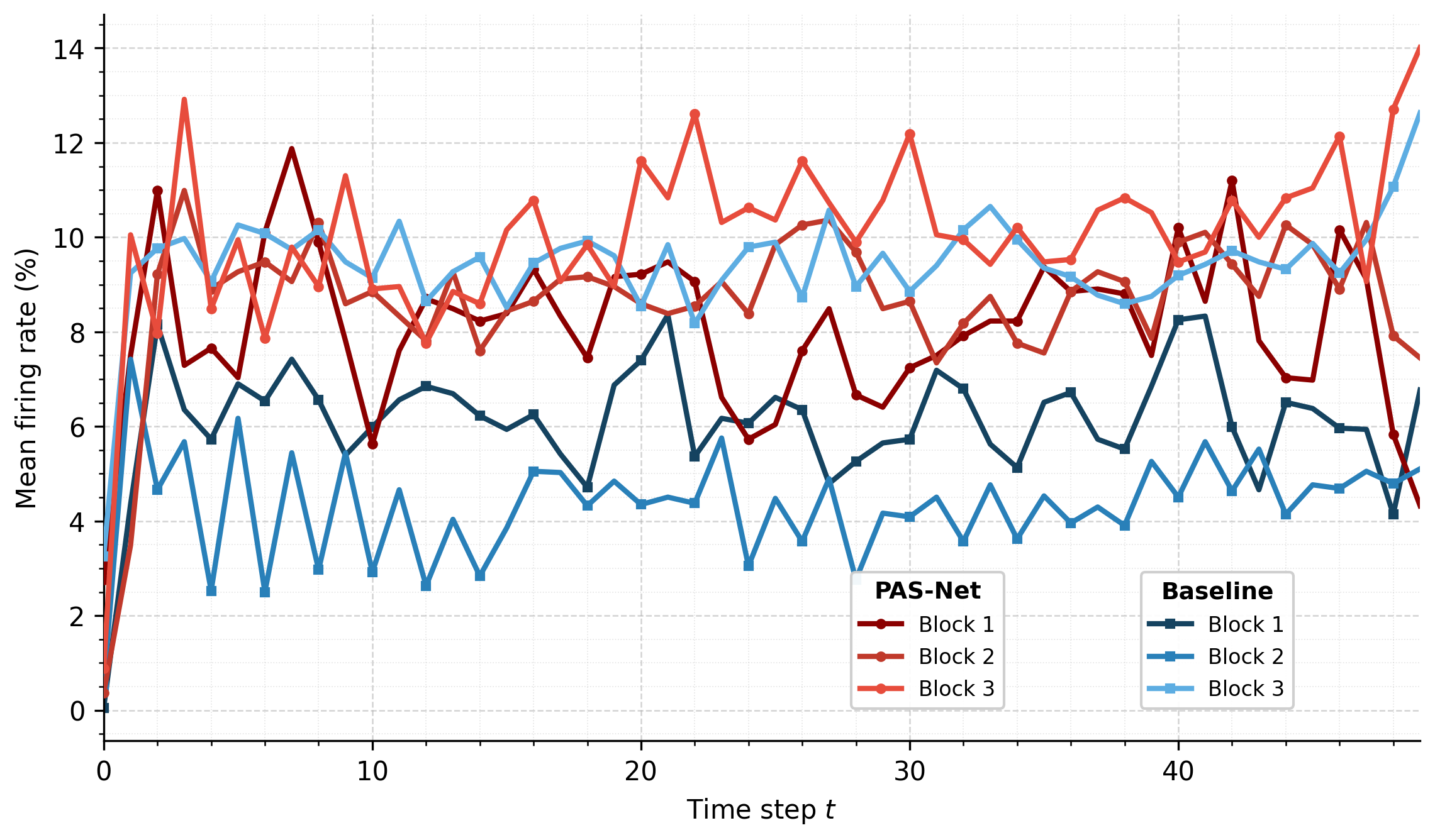}
        \caption{Running sequence}
        \label{fig:valve_running}
    \end{subfigure}
    
    \caption{Temporal micro-dissection of layer-wise firing rates across consecutive time steps ($t$). PAS-Net (red hues) exhibits a highly oscillatory, ``peak-and-valley'' rhythm tightly coupled with actual biomechanical phases (e.g., foot strikes), whereas the static Baseline (blue hues) fires in a blind, decoupled manner.}
    \label{fig:smart_valve_dynamics}
\end{figure*}

To quantitatively validate the ``computation-on-demand'' capability and the high dynamic range of our causal neuromodulator, we conduct a micro-dissection of the temporal firing dynamics. Figure~\ref{fig:smart_valve_dynamics} visualizes the step-by-step mean firing rates of the internal PAS-Blocks during real, continuous continuous IMU sequences (Walking and Running) from the PAMAP2 dataset. We compare our full PAS-Net (equipped with True Dynamic Threshold LIFs) against a degraded Baseline (using standard static LIFs with $V_{th}=1.0$).

As illustrated in Figure~\ref{fig:smart_valve_dynamics}, an initial, counter-intuitive observation is that PAS-Net occasionally exhibits localized firing rates that are higher than the static baseline (e.g., Block 1 peaking at $\sim 13\%$). However, this phenomenon explicitly elucidates the \textit{Sparsity Paradox} in neuromorphic HAR. 

Unlike the static baseline (blue lines), which maintains a relatively flat and blind firing pattern irrespective of the user's actual movement phase, PAS-Net (red lines) breathes with the physical kinematics. The dynamic LIF neurons act as intelligent semantic valves characterized by extreme temporal variance. During intense kinetic phases such as heel strikes (the sharp peaks), the causal gate actively lowers the membrane threshold, intentionally causing a spike burst to precisely ingest critical structural semantics and preserve classification accuracy. Conversely, during the swing or idle phases (the deep valleys), the threshold is instantly elevated, suppressing the firing rate significantly. 

This rhythmic, high-dynamic-range oscillation proves that PAS-Net does not pursue a naive, rigid sparsity that destroys information (spike vanishing). Instead, it executes true computation-on-demand. By dynamically expanding its cognitive effort only when physically necessary, PAS-Net filters out background noise at the architectural bottleneck, enabling the entire spatiotemporal system to achieve state-of-the-art accuracy while maintaining an ultra-low global energy footprint.
\section{Micro-Dissection of Spatiotemporal Firing Dynamics and Semantic Enrichment}
\label{app:spatiotemporal_matrix}

To quantitatively dissect the internal information routing and the unique ``computation-on-demand'' capability of PAS-Net, we conduct a micro-level spatiotemporal analysis. Table~\ref{tab:spatiotemporal_matrix} presents the layer-wise average firing rate across 8 uniform temporal segments ($T'_1$ to $T'_8$) of a single highly dynamic ``Running'' sequence from the PAMAP2 dataset.

\begin{table}[htbp]
\centering
\caption{Spatiotemporal Firing Matrix on the PAMAP2 dataset. The table tracks the layer-wise average firing rate (\%) across 8 uniform temporal segments ($T'_1$ to $T'_8$) of a downsampled sequence. Bold values highlight the ``Semantic Spike Enrichment'' phenomenon in the deepest temporal layers during critical kinematic phases.}
\label{tab:spatiotemporal_matrix}
\resizebox{0.95\linewidth}{!}{
\begin{tabular}{lcccccccc}
\toprule
\textbf{Layer Name} & \textbf{$T'_1$} & \textbf{$T'_2$} & \textbf{$T'_3$} & \textbf{$T'_4$} & \textbf{$T'_5$} & \textbf{$T'_6$} & \textbf{$T'_7$} & \textbf{$T'_8$} \\
\midrule
Stem & 7.40\% & 9.45\% & 7.42\% & 7.42\% & 8.07\% & 5.86\% & 8.55\% & 8.16\% \\
\midrule
Block1\_Topo & 7.22\% & 10.60\% & 7.34\% & 6.81\% & 8.72\% & 7.86\% & 8.03\% & 7.55\% \\
Block1\_Dynamic & 8.33\% & 9.97\% & 8.29\% & 6.90\% & 7.55\% & 7.73\% & 8.03\% & 7.99\% \\
Block1\_MLP1 & 6.03\% & 9.51\% & 7.07\% & 7.31\% & 6.81\% & 6.81\% & 7.83\% & 7.62\% \\
Block1\_MLP2 & 8.30\% & 12.83\% & 8.03\% & 8.77\% & 9.51\% & 7.94\% & 9.94\% & 11.20\% \\
\midrule
Block2\_Topo & 6.40\% & 12.05\% & 8.59\% & 7.38\% & 10.63\% & 8.29\% & 9.59\% & 9.38\% \\
Block2\_Dynamic & 7.07\% & 9.97\% & 9.77\% & 8.29\% & 10.07\% & 7.99\% & 9.11\% & 9.16\% \\
Block2\_MLP1 & 5.13\% & 8.98\% & 10.68\% & 9.03\% & 7.55\% & 7.62\% & 8.16\% & 8.72\% \\
Block2\_MLP2 & 3.09\% & 11.31\% & 17.14\% & 14.89\% & 12.85\% & 10.68\% & 12.02\% & 11.94\% \\
\midrule
Block3\_Topo & 7.33\% & 10.12\% & 5.90\% & 7.64\% & 10.16\% & 7.55\% & 8.90\% & 12.07\% \\
Block3\_Dynamic & 7.33\% & 11.50\% & 9.77\% & 11.98\% & 11.72\% & 11.02\% & 10.29\% & 12.41\% \\
Block3\_MLP1 & 3.98\% & 7.68\% & 9.72\% & 9.94\% & 11.28\% & 12.80\% & 14.15\% & 13.67\% \\
Block3\_MLP2 & 3.31\% & 11.57\% & \textbf{16.88\%} & \textbf{20.88\%} & \textbf{22.31\%} & \textbf{25.26\%} & \textbf{25.26\%} & \textbf{23.78\%} \\
\bottomrule
\end{tabular}
}
\end{table}

As illustrated in Table~\ref{tab:spatiotemporal_matrix}, PAS-Net exhibits a highly counter-intuitive but profoundly logical spatiotemporal firing distribution. Rather than maintaining an extremely rigid sparsity across the entire deep hierarchy, the network implements a dynamic \textbf{Semantic Spike Enrichment} mechanism.

In the shallow layers (e.g., \textit{Stem}), the firing rate remains remarkably stable and sparse ($\sim 7\% - 9\%$). This indicates that the perceptual frontend acts as a rigorous noise filter, successfully rejecting high-frequency sensor perturbations. However, as information propagates into the deepest temporal mixing layers (e.g., \textit{Block3\_MLP2}), the network intentionally amplifies its firing rate up to \textbf{25.26\%} during critical kinematic phases (from $T'_6$ to $T'_8$).

This structural ``inverted sparsity'' is an intentional architectural necessity. To accurately classify highly complex, non-stationary locomotion, the penultimate layers must assemble a rich, high-dimensional semantic representation just before the classification head. Instead of firing blindly at all times, PAS-Net intelligently allocates its \textit{spike energy budget}: it suppresses redundant firings in shallow layers and idle temporal segments, saving enough energy to deliberately unleash a dense burst of ``semantic spikes'' exactly when discriminative evidence is paramount. Consequently, PAS-Net guarantees high-confidence predictions (enabling sub-second early exits) while maintaining an ultra-low global energy profile.
\end{document}